\pdfoutput=1
\documentclass{article}
\usepackage{graphicx}
\usepackage{grffile}

\usepackage[final,nonatbib]{neurips_2022}

\usepackage{amsmath}
\usepackage{amssymb}
\usepackage{longtable}
\usepackage{mathtools}
\usepackage{amsthm}
\usepackage{pifont}
\usepackage{thmtools,thm-restate}
\usepackage{multirow}
\usepackage{multicol}
\usepackage{enumitem}
\usepackage{times}
\usepackage{bbm,bm}
\usepackage{soul}
\usepackage{url}
\usepackage[small]{caption}
\usepackage[utf8]{inputenc}
\usepackage{subfigure}
\usepackage{float}
\usepackage{tabularx}
\usepackage{pifont}
\usepackage{diagbox}
\usepackage{colortbl}
\usepackage{xcolor}
\usepackage{wrapfig}
\usepackage[titletoc]{appendix}
\usepackage{hyperref}
\usepackage{listings}
\usepackage{makecell}
\usepackage{algpseudocode}
\usepackage{algorithm}
\usepackage{algorithmicx}
\usepackage{booktabs}

\usepackage[capitalize,noabbrev]{cleveref}

\theoremstyle{plain}

\theoremstyle{definition}

\theoremstyle{remark}

\usepackage[textsize=tiny]{todonotes}

\usepackage{amsmath,amsfonts,bm}

\def\eqref#1{equation~\ref{#1}}

\def\1{\bm{1}}

\DeclareMathAlphabet{\mathsfit}{\encodingdefault}{\sfdefault}{m}{sl}
\SetMathAlphabet{\mathsfit}{bold}{\encodingdefault}{\sfdefault}{bx}{n}

\newcommand{\fig}[1]{Fig.~\ref{#1}}
\newcommand{\eq}[1]{Eq.~(\ref{#1})}
\newcommand{\tb}[1]{Tab.~\ref{#1}}
\newcommand{\se}[1]{Section~\ref{#1}}
\newcommand{\ap}[1]{Appendix~\ref{#1}}

\newcommand{\alg}[1]{Algo.~\ref{#1}}

\newcommand{\bbE}{\ensuremath{\mathbb{E}}} 
\newcommand{\caA}{\ensuremath{\mathcal{A}}} 

\newcommand{\caT}{\ensuremath{\mathcal{T}}} 
 
\newcommand{\caH}{\ensuremath{\mathcal{H}}} 
\newcommand{\caI}{\ensuremath{\mathcal{I}}} 
\newcommand{\caP}{\ensuremath{\mathcal{P}}} 
\newcommand{\caZ}{\ensuremath{\mathcal{Z}}}

\title{PerfectDou: Dominating DouDizhu with\\ Perfect Information Distillation}

\author{%
    Guan Yang$^{1}$\thanks{Equal contribution. Yang is responsible for the basic idea, system design and implementation details; Liu mainly contributes to the methodology, writing and experimental design. \dag Corresponding author. Project page at \url{https://github.com/Netease-Games-AI-Lab-Guangzhou/PerfectDou/}.}~~~ Minghuan Liu$^{2*}$~~~Weijun Hong$^{1}$~~~\\
    \textbf{Weinan Zhang$^{2}$~~~Fei Fang$^{3}$~~~Guangjun Zeng$^{1}$~~~Yue Lin$^{1\dag}$}\\
    {$^1$ NetEase Games AI Lab, $^2$ Shanghai Jiao Tong University, $^3$ Carnegie Mellon University}\\
  \texttt{\{yangguan,gzzengguanjun,gzlinyue\}@corp.netease.com},\\
  \texttt{\{minghuanliu,wnzhang\}@sjtu.edu.cn},~~~ \texttt{feif@cs.cmu.edu}
}

\begin{document}

\maketitle

\begin{abstract}

As a challenging multi-player card game, DouDizhu has recently drawn much attention for analyzing competition and collaboration in imperfect-information games. In this paper, we propose PerfectDou, a state-of-the-art DouDizhu AI system that dominates the game, in an actor-critic framework with a proposed technique named perfect information distillation. 
In detail, we adopt a perfect-training-imperfect-execution framework that allows the agents to utilize the global information to guide the training of the policies as if it is a perfect information game and the trained policies can be used to play the imperfect information game during the actual gameplay. To this end, we characterize card and game features for DouDizhu to represent the perfect and imperfect information.
To train our system, we adopt proximal policy optimization with generalized advantage estimation in a parallel training paradigm. In experiments we show how and why PerfectDou beats all existing AI programs, and achieves state-of-the-art performance.

\end{abstract}

\section{Introduction}

With the fast development of Reinforcement Learning (RL), game AI has achieved great success in many types of games, including board games (e.g., Go~\cite{silver2017masteringgo}, chess~\cite{silver2017mastering}), card games (e.g., Texas Hold'em~\cite{brown2018superhuman}, Mahjong~\cite{li2020suphx}), and video games (e.g., Starcraft~\cite{vinyals2019grandmaster}, Dota~\cite{berner2019dota}). 
As one of the most popular card games in China, DouDizhu has not been studied in depth until very recently. 
In perfect-information games such as Go, agent can observe all the events occurred previously including initial hand of each agent and all agents' actions. In contrast, DouDizhu is an imperfect-information game with special structure, and an agent does not know other agents' initial hands but can observe all agents' actions. 
One challenge in DouDizhu is that it is a three-player game with both competition and collaboration: the two \textit{Peasant} players need to cooperate as a team to compete with the third \textit{Landlord} player.
In addition, DouDizhu has a large action space that is hard to be abstracted for search-based methods~\cite{zhadouzero21}.

Although various methods have been proposed for tackling these challenges~\cite{you2019combinational,jiang2019deltadou}, they are either computationally expensive or far from optimal, and highly rely on abstractions with human knowledge~\cite{zhadouzero21}. 
Recently, Zha et al.~\cite{zhadouzero21} proposed DouZero, which applies simple Deep Monte-Carlo (DMC) method to learn the value function with pre-designed features and reward function. DouZero is regarded as the state-of-the-art (SoTA) AI system of DouDizhu for its superior performance and training efficiency compared with previous works.


Unfortunately, we find DouZero still has severe limitations in many battle scenarios, which will be characterized in \se{sec:case-study}. To establish a stronger and more robust DouDizhu bot, in this paper, we present a new AI system named PerfectDou, and show that it leads to significantly better performance than existing AI systems including DouZero. The name of our program follows the key technique we use -- perfect information distillation. The proposed technique utilizes a Perfect-Training-Imperfect-Execution (PTIE) framework, a variant of the popular Centralized-Training-Decentralized-Execution (CTDE) paradigm in multi-agent RL literature~\cite{FoersterAFW16,LoweMulti17}. Namely, we feed perfect-information to the agent in the training phase to guide the training of the policy, and only imperfect-information can be used when deploying the learned policy for actual game play. Correspondingly, we further design the card and game features to represent the perfect and imperfect information.
To train PerfectDou, we utilize Proximal Policy Optimization (PPO) \cite{schulman2017proximal} with Generalized Advantage Estimation (GAE) \cite{schulman2015high} by self-play in a distributed training system.

In experiments, we show PerfectDou beats all the existing DouDizhu AI systems and achieves the SoTA performance in a 10k-decks tournament; moreover, PerfectDou is the most training efficient, such that the number of samples required is an order of magnitude lower than the previous SoTA method; for application usage, PerfectDou can be deployed in online game environment due to its low inference time.


    

\section{Preliminaries}
\label{sec:pre}

\paragraph{Imperfect-Information Extensive-Form Games.}
An imperfect-information extensive-form (or tree-form) game can be described as a tuple $G=(\caP, \caH, \caZ, \caA, \caT, \chi, \rho, r, \caI)$, where $\mathcal{P}$ denotes a finite set of \textit{players}, $\caA$ is a finite set of actions, and $\caH$ is a finite set of \textit{nodes} at which players can take actions and are similar to states in an RL problem. At a node $h\in\caH$, $\chi:\caH\rightarrow 2^{\caA}$ is the action function that assigns to each node $h\in\caH$ a set of possible actions, and $\rho:\caH\rightarrow \caP$ represents the unique acting player. An action $a \in A(h)$ that leads from $h$ to $h'$ is denoted by the successor function $\caT:\caH\times\caA\rightarrow\caH$ as $h'= \caT(h,a)$.
$\caZ \subseteq \caH$ are the sets of terminal nodes for which no actions are available. For each player $p \in \caP$, there is a reward function $r_p\in r={r_1,r_2,\ldots, r_{|\caP|}}: \caZ\rightarrow \mathbb{R}$.
Furthermore, $\caI=\{\caI_p|p\in\caP\}$ describes the information sets (infosets) in the game where $\caI_p$ is a partition of all the nodes with acting player $p$. If two nodes $h,h'$ belong to the same infoset $I$ of player $p$, i.e., $h,h'\in I \in \caI_p$, these two nodes are indistinguishable to $p$ and will share the same action set. We use $I(h)$ to denote the infoset of node $h$.
Upon a certain infoset, a policy (or a behavior strategy) $\pi_p$ for player $p$ describes which action the player would take at each infoset. A policy can be stochastic, and we use $\pi_p(I)$ to denote the probability vector over player $p$'s available actions at infoset $I$. With a slight abuse of notation, we use $\pi_p(h)$ to denote the stochastic action player $p$ will take at node $h$. For two nodes $h, h'$ that belong to the same infoset $I$, it is clear that $\pi_p(h)=\pi_p(h')=\pi_p(I)$.
Therefore, the objective for each player $p$ is to maximize its own total expected return at the end of the game: $R_{p} \triangleq \bbE_{Z\sim\pi}[r_{p}(Z)], Z \in \caZ$.


\paragraph{The Three-Players DouDizhu Game.}

DouDizhu (a.k.a. Fight the Landload) is a three-player card game that is popular in China and is played by hundreds of  millions of people. Among the three players, two of them are called the \textit{Peasants}, and they need to cooperate as a team to compete against another player called the \textit{Landlord}. 
The standard game consists of two phases, bidding and cardplay. The bidding phase designates the roles to the players and deals leftover cards to the \textit{Landlord}. In the cardplay phase, the three players play cards in turn in clock-wise order. Within a game episode, there are several \textit{rounds}, and each begins with one player showing a legal combination of cards (solo, pair, etc.). The subsequent players must either choose to pass or beat the previous hand by playing a more superior combination of cards, usually in the same category. The round continues until two consecutive players choose to pass and the player who played the last hand initiates to the next round. DouDizhu is in the genre of shedding where the player wins by emptying his's hand, or loses vice versa. Therefore, in this game, the suit does not matter but the rank does. The score of a game is calculated as the base score multiplied by a multiplier determined by specialized categories of cards (\ap{ap:score} shows the details). 
In this paper, we only consider the cardplay phase, which can be formulated as an imperfect-information game. More detailed information about the game can be referred to \cite{zhadouzero21}.

The key challenges of DouDizhu include how the \textit{Peasants} work as a team to beat the \textit{Landlord} with card number advantage using only imperfect information. For example, one \textit{Peasant} can try to help his teammate to win by always trying the best to beat the \textit{Landlord}'s cards and play cards in a category where the teammate has an advantage.
In addition, the action space of DouDizhu is particularly large, and there are 27,472 possible combinations of cards that can be played in total with hundreds of legal actions in a hand. Furthermore, the action space cannot be easily abstracted since improperly playing a card may break other potential card combinations in the following rounds and lead to losing the game.
\section{Methodology}
\label{sec:method}

In this section, we first introduce how perfect-training-imperfect-execution works for a general imperfect-information game. Then, we formulate the DouDizhu game as an imperfect-information game to solve. 

\subsection{Perfect Information Distillation}
\label{sec:pid}

\begin{wrapfigure}{r}{0.37\textwidth}
\centering
\vspace{-10pt}
\includegraphics[height=0.95\linewidth]{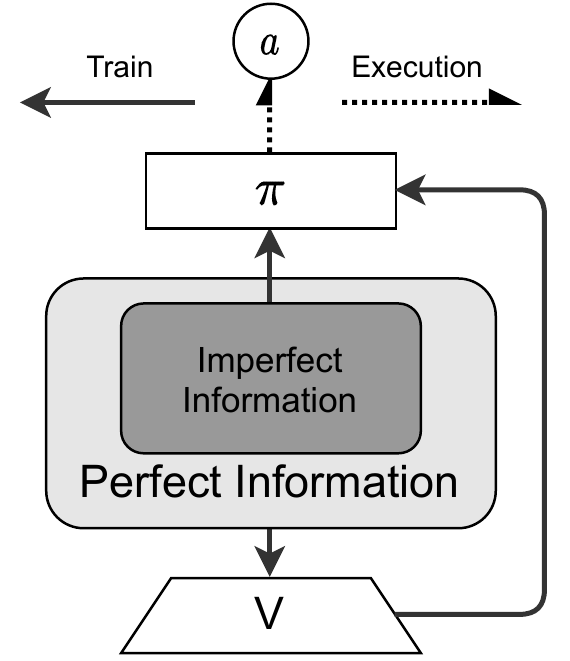}
\vspace{-4pt}
\caption{Overview of perfect information distillation within a perfect-training-imperfect-execution framework. The value network takes additional information (such as other players' cards in Poker games) as input, while the policy network does not.}
\label{fig:overview}
\vspace{-10pt}
\end{wrapfigure}
In card games as DouDizhu, the imperfect-information property comes from the fact that players do not show their hand cards to the others. And therefore the critical challenge for each player is to deal with the indistinguishable nodes from the same infoset. For such games, consider we can construct a strategically identical perfect-information game and allow one player to observe distinguishable nodes, then, the decisions at each node can rely on the global information and he may have more chances to win the game, like owning a cheating plug-in. This motivates us to utilize the distinguishable nodes for training the agents of imperfect-information games, and therefore we propose the technique of perfect information distillation. 

In general, the perfect information distillation is a framework trained in perfect-training-imperfect-execution (PTIE) paradigm, a variant of centralized-training-decentralized-execution (CTDE)~\cite{FoersterAFW16,LoweMulti17}, as illustrated in \fig{fig:overview}.
Particularly, CTDE constructs the value function with all agents' observations and actions for general multi-agent tasks. By comparison, our proposed PTIE is designed for imperfect-information games where additional perfect information is introduced in the training stage.
In this work, we consider to initiate PTIE with actor-critic~\cite{sutton2018reinforcement} (but PTIE is not limited to actor-critic), which is a template of policy gradient (PG) methods, proposed in the RL literature towards maximizing the expected reward of the policy function through PG with a value function:
\begin{equation}
    \nabla_{\theta_p}J = \bbE_{\pi}[\nabla_{\theta_p}\log{\pi_{\theta_p}(a|s)}Q_{\pi}(s, a)]~,
\end{equation}
where $s$ denotes a state in an RL problem, $Q$ is the state-action value function learned by a function approximator, usually called the critic. Notice that the critic is playing the role of evaluating how good an action is taken at a specific situation, but only at the training time. When the agent is deployed into inference, only the policy $\pi$ can be used to inferring feasible actions. Therefore, for imperfect-information games, we can provide additional information about the exact node the player is in to train the critic with self-play, as long as the actor does not take such information for decision making. Intuitively, we are distilling the perfect information into the imperfect policy.

Formally, for each node $h$, we construct a distinguishable node $D(h)$ for the strategically identical perfect-information game. Then, we define the value function at $D(h)$, $V_{\pi_p}(D(h))=\bbE_{a\sim \pi_p, h^0=h} [Q_{\pi_p}(D(h),a)]=\bbE_{Z | \pi_p, h^0=h} [r_p(Z)]$ as the expected value of distinguishable nodes. 
In the sequel, we propose a simple extension of actor-critic policy gradient considering parameterized policy $\pi_{\theta_p}$ for each player $p$:
\begin{equation}\label{eq:ptie-pg}
    \nabla_{\theta_p}J = \bbE_{\pi_p}[\nabla_{\theta_p}\log{\pi_{\theta_p}(a|h)}Q_{\pi_p}(D(h), a)]~.
\end{equation}  
In practice, we use a policy network (actor) to represent the policy $\pi_p$ for each player, which takes as input a vector describing the representation at an indistinguishable node $h$ that the player can observe during the game. For estimating the critic, a value network is utilized, which takes representation of the global information at the distinguishable node $D(h)$. 
In other words, the value network takes additional information as inputs (such as other players' cards in Poker games) or targets (such as immediate rewards computed upon others' hands), while the policy network does not. 
During the training, the value function updates the values for all distinguishable nodes; then, it trains the policy on every node on the same infoset from sampled data, which implicitly gives an expected value estimation on each infoset. In practice, the generalization ability of neural networks enables the policy to find a better solution, which is also the advantage for using the proposed PTIE framework.

PTIE is a general way for training imperfect-information game agents using RL. 
In addition, the optimality point of independent RL in multi-agent learning is exactly one of an Nash equilibria (NE) and it has been proved that training RL algorithms with self-play can converge to an NE in two-player cases (although no convergence guarantees for three-player games)~\cite{heinrich2016deep,lanctot2017unified}, which may provide insights why PTIE works. 
We expect that with PTIE, players can leverage the perfect information during inference to derive coordination and strategic policies. 
In experiments, we show that this allows PerfectDou to cooperate with each other (as \textit{Peasants}) or compete against the team (as the \textit{Landlord}).




\subsection{DouDizhu as An Imperfect-Information Game}

As is mentioned in \se{sec:pre}, the cardplay phase of DouDizhu can be regarded as an imperfect-information game with three players. 
At each node $h$, its infoset $I_p$ contains all combinations of the other's invisible handcards. Then at each level of the game tree, the three players take clockwise turns to choose an available action with policies $\pi_p$ depending on the infoset $I_p(h)$ of the current node $h$ with the reward function $r_p$. The path from the root of the game tree to a node $h$ contains the initial hand of all players and all the historical moves of all players. The reward functions at leaf nodes are set to be the score the players win or lose at the end of the game.

\section{PerfectDou System Design}
\label{sec:system}

\begin{wrapfigure}{r}{0.52\textwidth}
\centering
\vspace{-10pt}
\includegraphics[width=0.99\linewidth]{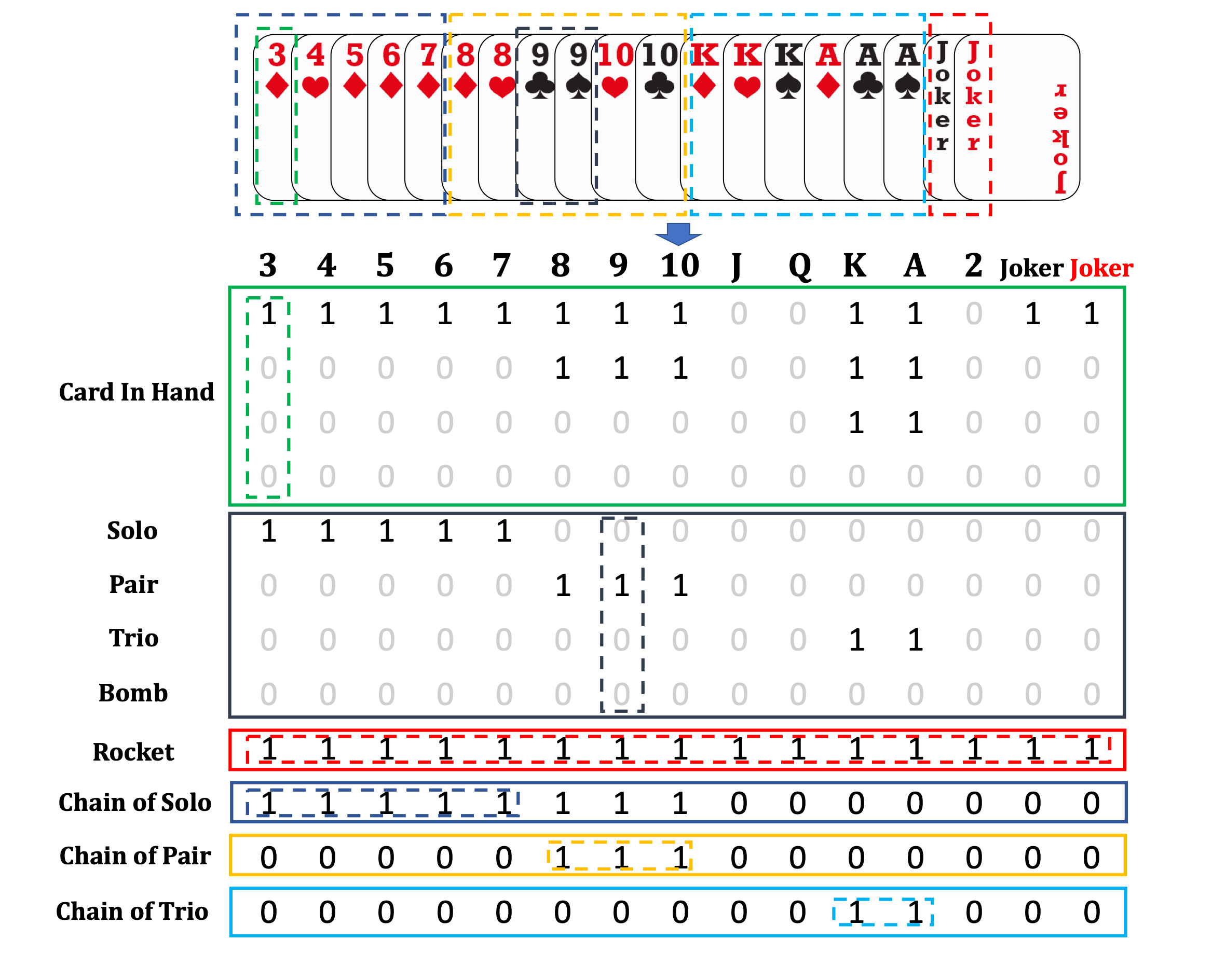}
\vspace{-7pt}
\caption{Card representation matrix. Columns stand for 15 different card ranks and rows stand for correspondingly designed features. The first 4 rows are the same as Zha et al.~\cite{zhadouzero21}, and the last 8 rows are additional design for encoding the legal combination of cards.}
\label{fig:card-matrix}
\vspace{-30pt}
\end{wrapfigure}

In this section, we explain how we construct our PerfectDou system in detail, with the proposed perfect information distillation technique, and several novel components designed for DouDizhu that help it summit the game. Particularly, PTIE requires different representations as input layer for the policy and the value network by feeding the value function with perfect information (distinguishable nodes) and the policy with imperfect information (indistinguishable nodes). 

\subsection{Card Representation}

In our system, we encode each feasible card combination with a $12\times15$ matrix, as shown in \fig{fig:card-matrix}. Specifically, we first encode different ranks and numbers with a $4\times 15$ matrix, where the columns correspond to the 15 ranks (including jokers) and similar to Zha et al.~\cite{zhadouzero21}, the number of ones in the four rows of a single column represents the number of cards of that rank in the player's hand.
Different from Zha et al.~\cite{zhadouzero21}, we further propose to encode the legal combination of cards with the player's current hand, to help the agent realize the different property of various kinds of cards (see \ap{ap:terms}). The feature sizes of each part are shown in \ap{ap:card-representation}.

\begin{table}[t]
\begin{center}
\caption{Feature design of perfect-information (distinguishable nodes) and imperfect-information (indistinguishable nodes) for the game. Perfect features include all imperfect features.}
\label{tb:info-feature}
\begin{sc}
\resizebox{0.9\linewidth}{!}{
\begin{tabular}{c||c||c}
\toprule
         & \textbf{Feature Design} & \textbf{Size} \\
\midrule
  \multirow{15}{*}{\textbf{\makecell[c]{Imperfect\\ Feature\\}}} & current player's hand & $1 \times 12 \times 15$ \\
  & unplayed cards & $1 \times 12 \times 15$  \\
  & current player's played cards  & $1 \times 12 \times 15$ \\
  & previous player's played cards & $1 \times 12 \times 15$  \\
  & next player's played cards & $1 \times 12 \times 15$  \\
  & 3 additional bottom cards & $1 \times 12 \times 15$ \\
  & last 15 moves & $15 \times 12 \times 15$  \\
  & previous player's last move & $1 \times 12 \times 15$\\
  & next player's last move & $1 \times 12 \times 15$\\
  \cline{2-3}
  & {minimum play-out steps of hand cards} & {$1$} \\
  & {number of cards in current player's hand} & {$1$} \\
  & {number of cards in previous player's hand} & {$1$} \\
  & {number of cards in next player's hand} & {$1$} \\
  & {number of bombs} &  {$1$} \\
  & {flag of game control by current player} & {$1$} \\
  \hline
  \multirow{4}{*}{\textbf{\makecell[c]{Additional \\ Perfect\\ Feature\\}}} & {previous player's hand cards} & {$1 \times 12 \times 15$} \\
  & {next player's hand cards} & {$1 \times 12 \times 15$} \\
  \cline{2-3}
  & {minimum play-out steps of previous player's hand cards} & {$1$} \\
  & {minimum play-out steps of next player's hand cards} & {$1$} \\
\bottomrule
\end{tabular}
}
\end{sc}
\end{center}
\end{table}

\subsection{Node Representation}
\label{se:node-rep}
In the game of DouDizhu, the distinguishable node $D(h)$ should cover all players' hand cards at $h$, along with the game and player status. Therefore, we propose to represent $h$ with imperfect features and $D(h)$ with perfect feature designs, shown in \tb{tb:info-feature}. In detail, the imperfect features include a flatten matrix\footnote{Short for a matrix flattened to a one-dimensional vector.} of $23 \times 12 \times 15$ and a game state array of $6 \times 1$. On the contrary, the perfect features consist of a flatten card matrix of $25 \times 12 \times 15$ and a game state array of $8 \times 1$. Therefore, they are totally asymmetric, and the imperfect features are a subset of the perfect features.

\subsection{Network Structure and Action Representation}
\begin{wrapfigure}{r}{0.6\textwidth}
\centering
\vspace{-25pt}
\includegraphics[width=0.99\linewidth]{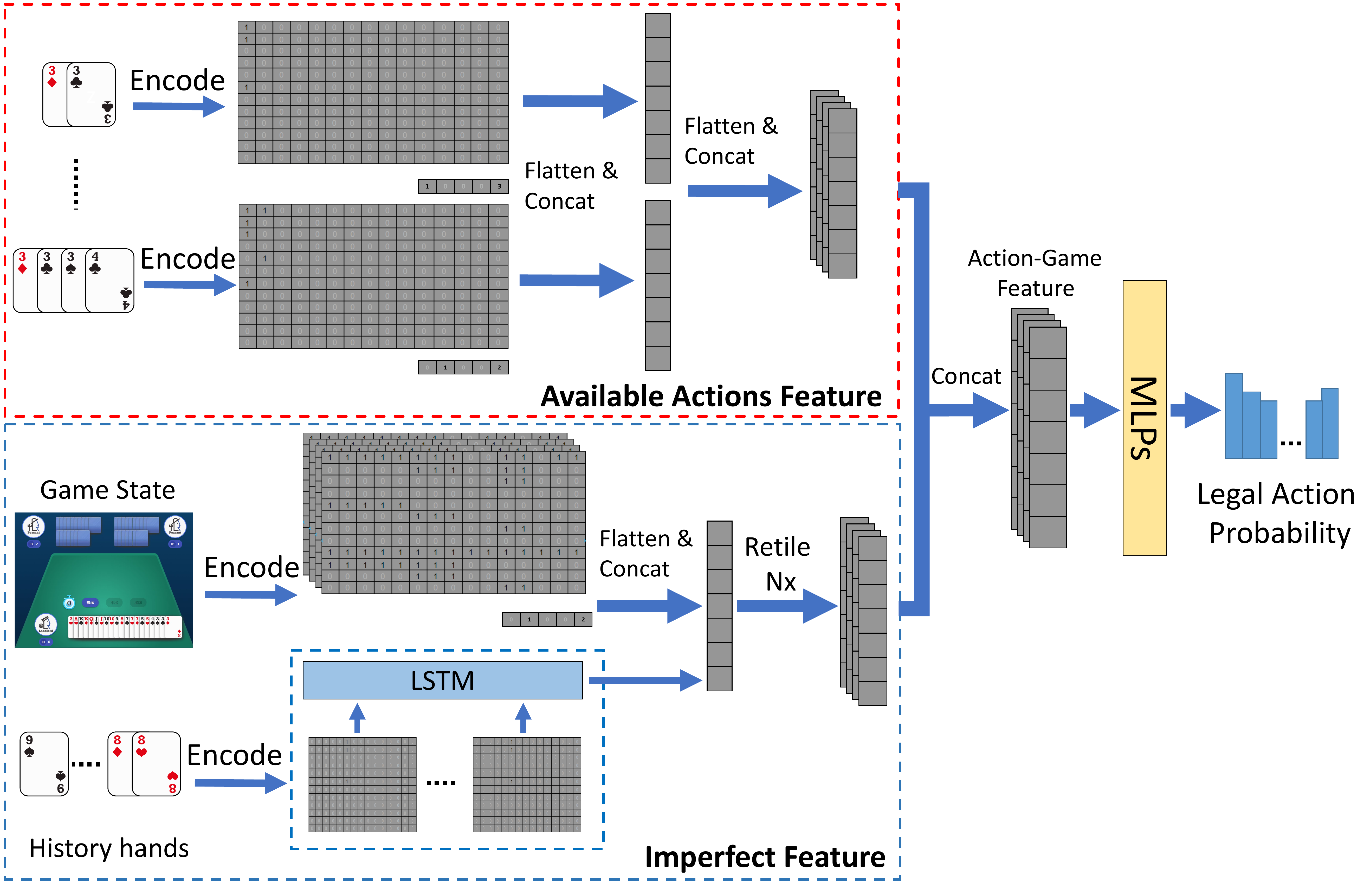}
\vspace{-6pt}
\caption{The policy network structure of PerfectDou system. The network predicts the action distribution given the current imperfect information of the game, including state information and available actions feature.}
\label{fig:action-net}
\vspace{-10pt}
\end{wrapfigure}

The PerfectDou system follows the general actor-critic design, and we take PPO~\cite{schulman2017proximal} with GAE~\cite{schulman2015high} as the learning algorithm. Slightly different from \eq{eq:ptie-pg}, PPO estimates the advantage $A_p=R_p-V_{\pi_p}$ as the critic instead of $Q_{\pi_p}$.
For value network, we use an MLP to handle encoded features (the detailed structure is shown in \ap{ap:value-structure}). As for the policy network, we first utilize an LSTM to encode all designed features; to encourage the agent to pay attention to specific card types, the proposed network structure will encode all the available actions into feature vectors, as depicted in \tb{tb:action-feature}. The output of the legal action probability is then computed with the action and game features, as illustrated in \fig{fig:action-net}. Formally, we concatenate the node representation $e_s$ with each action representation $e_{a^i}$ separately, and get the legal action distribution:
\begin{equation}
    p(a) = \text{softmax}(f([e_s,e_{a^i}]_{i=1}^N)~,
\end{equation}
where $a^i$ is the $i$-th action, $[\cdot]$ denotes the concatenation operation for $N$ available actions, and $f$ are layers of MLPs. This resembles the target attention mechanism in Ye et al.~\cite{ye2020mastering}.

\subsection{Perfect Reward Design}
\label{sec:reward}
If we only care about the result at the end of the game, the reward at leaf nodes is rather sparse; in addition, players can only estimate their advantage of winning the game using imperfect information during the game, which could be inaccurate and fluctuated. Thanks to PTIE, we are allowed to impose an oracle reward function for DouDizhu at each node to enhance the perfect information modeled by the value function.
In the training of PerfectDou, instead of estimating the advantage, we utilize an oracle\footnote{Implemented as a dynamic programming algorithm, see \ap{ap:implementation} for details.} for evaluating each player, particularly, the minimum steps needed to play out all cards, which can be treated as a simple estimation of the distance to win. The reward function is then defined as the advantage difference computed by the relative distance to win of the two camps in two consecutive timesteps.
Formally, at timestep $t$, the reward function is:
\begin{equation}\label{eq:reward}
r_t = \left\{
    \begin{aligned}
        -1.0 \times (\text{Adv}_{t}-\text{Adv}_{t-1})\times l,   & &   &\text{\textit{Landlord}} \\
        0.5 \times (\text{Adv}_{t}-\text{Adv}_{t-1}) \times l,   & &    &\text{\textit{Peasant}}
    \end{aligned}
\right.
\end{equation}
\begin{equation}\label{eq:advantage}
\text{Adv}_t = N_t^\text{\textit{Landlord}} - \min\left(N_t^\text{\textit{Peasant}1},N_t^\text{\textit{Peasant}2}\right)~,
\end{equation}
where $l$ is a scaling factor, and $N_t$ is the minimum steps to play out all cards at timestep $t$.

For instance, in a round, at timestamp $t$, the distance of the \textit{Landlord} to win is 5 and the distances of two \textit{Peasants} are 3 and 7, which means \textit{Peasants} have a larger advantage since the relative distance is 2 for \textit{Peasants} and -2 for the \textit{Landlord}. However, if the \textit{Landlord} plays a good hand such that both \textit{Peasants} can not suppress, the \textit{Landlord} will in result get a positive reward due to the decreased relative distance of the \textit{Landlord}, i.e., from 2 to 1. Correspondingly, the \textit{Peasants} would get a negative reward as their relative distances are getting larger. Such a reward function can encourage the cooperation between \textit{Peasants}, since the winning distance is defined by the minimum steps of both players.
In our implementation, the computation of the rewards is carried out after a round of the game, hence to promote training efficiency.

\subsection{Distributed Training Details}
\begin{figure}[htbp]
\centering
\vspace{-5pt}
\includegraphics[width=0.8\columnwidth]{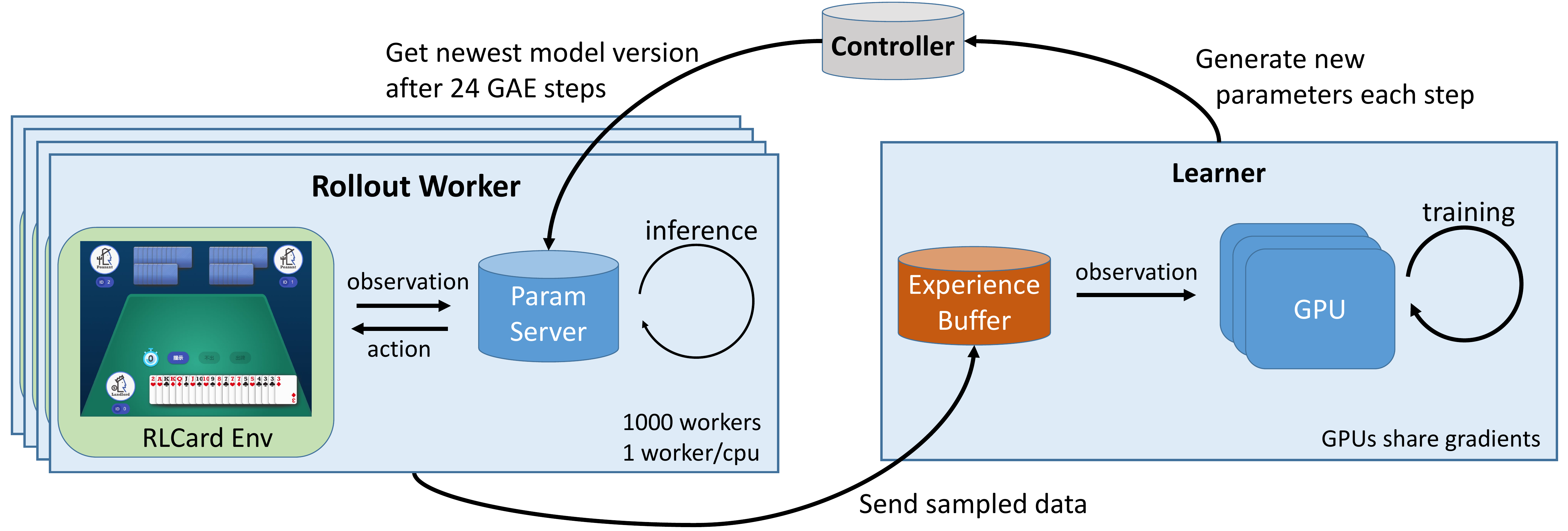}
\vspace{-3pt}
\caption{Illustration of the distributed training system.}
\label{fig:system}
\end{figure}

To further expedite the training procedure, we design a distributed training system represented in \fig{fig:system}. Specifically, the system contains a set of rollout workers for collecting the self-play experience data and sending it to a pool of GPUs; these GPUs asynchronously receive the data and store it into their local buffers. Then, each GPU learner randomly samples mini-batches from its own buffer and compute the gradient separately, which is then synchronously averaged across all GPUs and back propagated to update the neural networks. After each round of updating, new parameters are sent to every rollout worker. And each worker will load the latest model after 24 (8 for each player) steps sampling. Such a decoupled training-sampling structure will allow PerfectDou to be extended to large scale experiments. Our design of the distributed system borrows a lot from IMPALA~\cite{espeholt2018impala}, which also keeps a set of rollout workers to receive the updated model, interact with the environment and send back rollout trajectories to learners. The main difference is derived from the learning algorithm where we use PPO with GAE instead of actor-critic with V-trace~\cite{espeholt2018impala}. Moreover, we keep three different models for \textit{Landlord} and two \textit{Peasants} separately which are only updated by their own data against the latest opponent models.

\section{Related Work}

\paragraph{Imperfect-Information Games.}
Many popular card games are imperfect-information games and have attracted much attention. For instance, Li et al.~\cite{li2020suphx} worked on the four-player game Mahjong and proposed a distributed RL algorithm combined with techniques like global reward prediction, oracle guiding, and run-time policy adaptation to win against most top human players; in addition, Lerer et al.~\cite{lerer2020improving} adopted search-based and imitation methods to learn the playing policy for Hanabi. Iterative algorithms such as Counterfactual Regret Minimization (CFR) and its variants~\cite{zinkevich2007regret,moravvcik2017deepstack,brown2018superhuman} are also well-used for a fully competitive game, Hold'em Poker, whose action space is generally designed at the scale of tens (fold, call, check and kinds of bet) and the legal actions at each decision point are even less~\cite{moravvcik2017deepstack,zhao2022alphaholdem}. In addition, CFR does not proven to converged to an Nash equilibrium in games with more than two players~\cite{li2021survey}. Consequently, they can be hardly designed for DouDizhu due to the large action space with difficult abstraction and the mixed game property (both cooperative and competitive, although there are few success cases for such a setting~\cite{mazrooei2013automating}).


\paragraph{DouDizhu AI systems.}
Besides the recent SoTA work DouZero~\cite{zhadouzero21}, many researchers have made efforts on utilizing the power of RL into solving DouDizhu. However, simply applying RL algorithms such as DQN and A3C into the game can hardly make benefits~\cite{you2019combinational}. Therefore, You et al.~\cite{you2019combinational} proposed Combinational Q-Network (CQN) that reduces the action space by heuristics action decoupling; moreover, DeltaDou~\cite{jiang2019deltadou} utilized Monte-Carlo Tree Search (MCTS) for DouDizhu, along with Bayesian inference for the hidden information and a pre-trained kicker network for action abstraction. DeltaDou was also reported as reaching human-level performance. Zhang et al.~\cite{zhang2021combining} also relied on MCTS with predicting other players’ actions. Recently, Zhao et al.~\cite{zhao2022douzero+} similarly proposed to model opponents' actions and train the policy model based on the DouZero architecture, reaching limited improvements. Nevertheless, as shown in this paper, we can instead distill such perfect-information knowledge like opponents' hand cards to the policy in a perfect-training-imperfect-execution style and reach a better performance.
\section{Experiments}
\label{sec:exps}

\begin{table*}[tb]
    \centering
    \caption{DouDizhu tournaments for existing AI programs by playing 10k randomly generated decks. Player \texttt{A} outperforms \texttt{B} if WP is larger than 0.5 or ADP is larger than 0 (highlighted in \textbf{boldface}). The algorithms are ranked according to the number of the other algorithms that they beat. 
    We note that DouZero is the current SoTA DouDizhu bot and the \textcolor{gray}{gray} rows highlight the comparison. Numerical results except marked $*$ are directly borrowed from Zha et al.~\protect\cite{zhadouzero21}.}
    \label{tab:mainperformance}
    \vspace{-5pt}
    \small
    \vspace{-0pt}
    \setlength{\tabcolsep}{2.0pt}
    \resizebox{0.9\textwidth}{14mm}{
    \begin{tabular}{c||l|cc|cc|cc|cc|cc|cc}
    \toprule
     
\multirow{2}{*}{Rank} & \multirow{2}{*}{\diagbox [width=8em,trim=l] {\texttt{A}}{\texttt{B}}} & \multicolumn{2}{c|}{PerfectDou}& \multicolumn{2}{c|}{DouZero} & \multicolumn{2}{c|}{DeltaDou} & \multicolumn{2}{c|}{RHCP-v2} & \multicolumn{2}{c|}{CQN} &\multicolumn{2}{c}{Random}\\
\cline{3-14}
& & WP & ADP & WP & ADP & WP & ADP & WP & ADP & WP & ADP & WP & ADP \\
    \hline
    \midrule
     \rowcolor{gray!60} 1 & PerfectDou (Ours) & - & - & \textbf{0.543}$^*$ & \textbf{0.143}$^*$ & \textbf{0.584}$^*$ & \textbf{0.420}$^*$ & \textbf{0.543}$^*$ & \textbf{0.506}$^*$ & \textbf{0.862}$^*$ & \textbf{2.090}$^*$ & \textbf{0.994}$^*$ & \textbf{3.146}$^*$\\
     2 & DouZero (Paper) & - & - & - & - & \textbf{0.586} & \textbf{0.258} & \textbf{0.764} & \textbf{1.671} & \textbf{0.810} & \textbf{1.685} & \textbf{0.989} & \textbf{3.036}\\
     \rowcolor{gray!60} - & DouZero (Public) & 0.457$^*$ & -0.143$^*$ & - & - & \textbf{0.585}$^*$ & \textbf{0.253}$^*$ &  0.451$^*$ & \textbf{0.060}$^*$  & \textbf{0.828}$^*$ & \textbf{1.950}$^*$ & \textbf{0.986}$^*$ & \textbf{3.050}$^*$\\
     3 & DeltaDou & 0.416$^*$ & -0.420$^*$ & 0.414 & -0.258 & - & - & \textbf{0.691}$^*$  & \textbf{1.528}$^*$  & \textbf{0.784} & \textbf{1.534} & \textbf{0.992} & \textbf{3.099}\\
     4 & RHCP-v2 & 0.457$^*$ & -0.506$^*$ & \textbf{0.549}$^*$ & -0.060$^*$ & 0.309$^*$  & -1.423$^*$  & - & - & \textbf{0.770}$^*$ & \textbf{1.414}$^*$ & \textbf{0.990}$^*$ & \textbf{2.670}$^*$\\
     5 & CQN & 0.138$^*$ & -2.090$^*$ & 0.190 & -1.685 & 0.216 & -1.534 & 0.230$^*$ & -1.414 $^*$& - & - & \textbf{0.889} & \textbf{1.912}\\
     6 & Random & 0.006$^*$ & -3.146$^*$ & 0.011 & -3.036 & 0.008 & -3.099  & 0.010$^*$ & -2.670$^*$ & 0.111 & -1.912 & - & -\\
     \bottomrule
    \end{tabular}
    }
    \vspace{-7pt}
\end{table*}

We conduct comprehensive experiments to investigate the following research questions.
\textbf{RQ1}: How good is PerfectDou against SoTA DouDizhu AI? 
\textbf{RQ2}: What are the key ingredients of PerfectDou?
\textbf{RQ3}: How is the inference efficiency of PerfectDou?
To answer \textbf{RQ1}, we empirically evaluate the performance against existing DouDizhu programs. Regarding \textbf{RQ2}, we conduct ablation studies on key components in our design. And for \textbf{RQ3}, we calculate the average inference time for all algorithms involved. Finally, we conduct in-depth analysis, provide interesting case studies of PerfectDou. In the appendix, we report more results including a battle against skilled human players.

\subsection{Experimental Setups}
\paragraph{Baselines.}
We evaluate PerfectDou against the following algorithms under the open-source RLCard Environment~\cite{zha2019rlcard}: 1) \textbf{DouZero}~\cite{zhadouzero21}: A recent SoTA baseline method that had beaten every existing DouDizhu AI system using Deep Monte-Carlo algorithm.
2) \textbf{DeltaDou}~\cite{jiang2019deltadou}: An MCTS-based algorithm with Beyesian inference. It achieved comparable performance as human experts. 
3) \textbf{Combinational Q-Network (CQN)}~\cite{you2019combinational}: Based on card decomposition and Deep Q-Learning.
4) \textbf{Rule-Based Algorithms}: Including the open-source heuristic-based program \textbf{RHCP-v2}~\cite{jiang2019deltadou,zhadouzero21}, the rule model in RLCard and a \textbf{Random} program with uniform legal moves. 
For evaluation, we directly take their public (or provided) codes and pre-trained models.

\paragraph{Metrics.}
The performance of DouDizhu are mainly quantified following the same metrics in previous researches~\cite{jiang2019deltadou,zhadouzero21}. Specifically, given two algorithms \texttt{A} against \texttt{B}, we calculate: 
1) \textbf{WP}~(Winning Percentage): The proportion of winning by \texttt{A} in a number of games.
2) \textbf{ADP}~(Average Difference in Points):
The per-game averaged difference of scores between \texttt{A} and \texttt{B}. In other words, positive ADP means gaining scores while the negative represents losing it. This is a more reasonable metric for evaluating DouDizhu AI systems, because in real games players are evaluated by the scores obtained instead of their winning rates, as further discussed in \ap{ap:score}.

In our experiments, we choose ADP as the basic reward for all experiments training PerfectDou,
 which is augmented with the proposed reward signal in \se{sec:reward} during the training stage.

\subsection{Comparative Evaluations}
\label{sec:comp_eval}

We conduct a tournament to demonstrate the advantage of our PerfectDou, where each pair of the algorithms play 10,000 decks, shown in \tb{tab:mainperformance} (\textbf{RQ1}). Since the bidding performance in each algorithm varies and poor bidding would affect game results significantly, for fair comparison, we omit the bidding phase and focus on the phase of cardplay. In detail, all games are randomly generated and each game would be played two times, i.e., each competing algorithm is assigned as \textit{Landlord} or \textit{Peasant} once. We use WP and ADP as the basic reward respectively for comparing over these two metrics for all evaluating methods.

Overall, PerfectDou dominates the leaderboard by beating all the existing AI programs, no matter rule-based or learning based algorithms, with significant advantage on both WP and ADP. Specifically, as noted that DouDizhu has a large variance where the initial hand cards can seriously determine the advantage of the game; even though, PerfectDou still consistently outperforms the current SoTA baseline -- DouZero. However, we find that PerfectDou is worse than the result published in DouZero paper~\cite{zhadouzero21} when competing against RHCP. To verify this problem, we test the public model of DouZero (denoted as DouZero (Public) with grey color). To our surprise, its performance can match most of the reported results in their paper except the one against RHCP, where it only takes a WP of 0.452 lower than 0.5, indicating that the public model of DouZero can not beat RHCP as suggested in the original paper, and in fact PerfectDou is the better one. 

It is also observed that some competition outcome has a high WP and a negative ADP. A potential reason can be explained as such agents are reckless to play out the bigger cards without considering the left hand, leading to winning many games of low score, but losing high score in the other games when their hand cards are not good enough.
From our statistics of online human matches, the WP of winner is usually in a range of $0.52\sim0.55$ when the player tries to maximize its ADP.


  \begin{wraptable}{r}{0.55\textwidth}
    \vspace{-10pt}
  \caption{Training efficiency comparison over 100k decks.}
    \label{tab:efficiency}
    \small
    \resizebox{0.99\linewidth}{10mm}{
    \begin{tabular}{l|cc|cc}
    \toprule
    \multirow{2}{*}{\diagbox [width=9em,trim=l] {\texttt{A}}{\texttt{B}}} & \multicolumn{2}{c|}{DouZero ($\sim$1e9)}& \multicolumn{2}{c}{DouZero ($\sim$1e10)}\\
    \cline{2-5}
    & WP & ADP & WP & ADP  \\
    \hline
    \midrule
     PerfectDou (2.5e9) & - & - & 0.541 & 0.130\\
     PerfectDou (1e9) & 0.732 & 1.270 & 0.524 & 0.014\\
     DouZero ($\sim$1e10) & 0.698 & 1.150 & - & -  \\
     \bottomrule
    \end{tabular}
    }
    \vspace{-6pt}
\end{wraptable}
We further reveal the sample efficiency of PerfectDou by comparing the competing performance w.r.t. different training steps. As shown in \tb{tab:efficiency}, we compare two versions of PerfectDou (1e9 and 2.5e9 steps) against two versions of DouZero (roughly 1e9 and 1e10 steps). From the tournament in \tb{tab:mainperformance}, we know the final version of PerfectDou (2.5e9 steps) outperforms the final version of DouZero ($\sim$1e10 steps). However, to our surprise, an early stage of PerfectDou is able to beat DouZero. With the same 1e9 training samples, PerfectDou wins DouZero with a large gap (WP of 0.732 and ADP of 1.270), which is even better than the 1e10 sample-trained DouZero. This indicates that PerfectDou is not only the best performance but also the most training efficient. The related training curves are shown in \ap{ap:curves}.

\subsection{Ablation Studies}
We want to further investigate the key to the success of our AI system (\textbf{RQ2}). Specifically, we would like to analyse how our design of the feature and the training framework help PerfectDou dominate the tournament of DouDizhu.
To this end, we evaluate different variants of PerfectDou and the previous SoTA AI system -- DouZero, including: a) \textit{ImperfectDouZero}\footnote{Note that DouZero cannot acquire any perfect-information feature since it will play in a cheating style if so.}: DouZero with our proposed imperfect-information features. b) \textit{ImperfectDou}: PerfectDou with only imperfect-information features as inputs for the value function. c) \textit{RewardlessDou}: PerfectDou without node reward.
d) \textit{Vanilla PPO}: Naive actor-critic training with imperfect-information features only and without additional reward.
\begin{wraptable}{r}{0.6\textwidth}
    \vspace{-10pt}
  \caption{Ablation studies over 100k decks.}
    \label{tab:ablation}
    \small
    \vspace{-0pt}
    \resizebox{0.99\linewidth}{9mm}{
    \begin{tabular}{l|cc|cc|cc}
    \toprule
     \multirow{2}{*}{\diagbox [width=10em,trim=l] {\texttt{A}}{\texttt{B}}} & \multicolumn{2}{c|}{DouZero ($\sim$1e9)}& \multicolumn{2}{c|}{DouZero ($\sim$1e10)} & \multicolumn{2}{c}{ImperfectDouZero ($\sim$1e9)}\\
    \cline{2-7}
    & WP & ADP & WP & ADP & WP & ADP  \\
    \hline
    \midrule
     PerfectDou (1e9) & 0.732 & 1.270 & 0.524 & 0.014 & 0.731 & 1.350 \\
     ImperfectDou (1e9) &0.717 & 1.180 & 0.486 & -0.057 & 0.723 & 1.320 \\
     RewardlessDou (1e9) &0.738 & 0.490 & 0.540 & -0.201 & 0.659 &  0.587\\
     Vanilla PPO(1e9) &0.509  & -0.307 & 0.346 & -0.709 & 0.433 & -0.023 \\
     \bottomrule
    \end{tabular}
    }
    \vspace{-3pt}
\end{wraptable}
The ablation experiments are designed as competitions among ImperfectDou, RewardlessDou and PerfectDou against DouZero for comparing the effectiveness of perfect information distillation and perfect intermediate reward separately; while the battle of ImperfectDouZero and DouZero against PerfectDou are designed for excluding the benefit from feature engineering.
Results for all comparisons are shown in \tb{tab:ablation}. 
Even with the imperfect features only, ImperfectDou can still easily beat DouZero with the same training steps; however, DouZero turns the tide with much more training data. Furthermore, our proposed node features seem not appropriate for DouZero to achieve a better results compared with its original design. Additionally, without the node reward, PerfectDou still beats DouZero with higher WP (in spite of sacrificing a lot of ADP), indicating the effectiveness of perfect reward in training, without which it would risk losing points to win one game. Finally, without both node reward and perfect feature design for the value function, vanilla PPO simply can not perform well. 
Therefore, we can conclude that our actor-critic based algorithm along with the PTIE training provides a high sample efficiency under our feature design, and the node reward benefits the rationality of our AI.

\subsection{Runtime Analysis}
\begin{wrapfigure}{r}{0.5\textwidth}
\centering
\vspace{-25pt}
\includegraphics[width=0.99\linewidth]{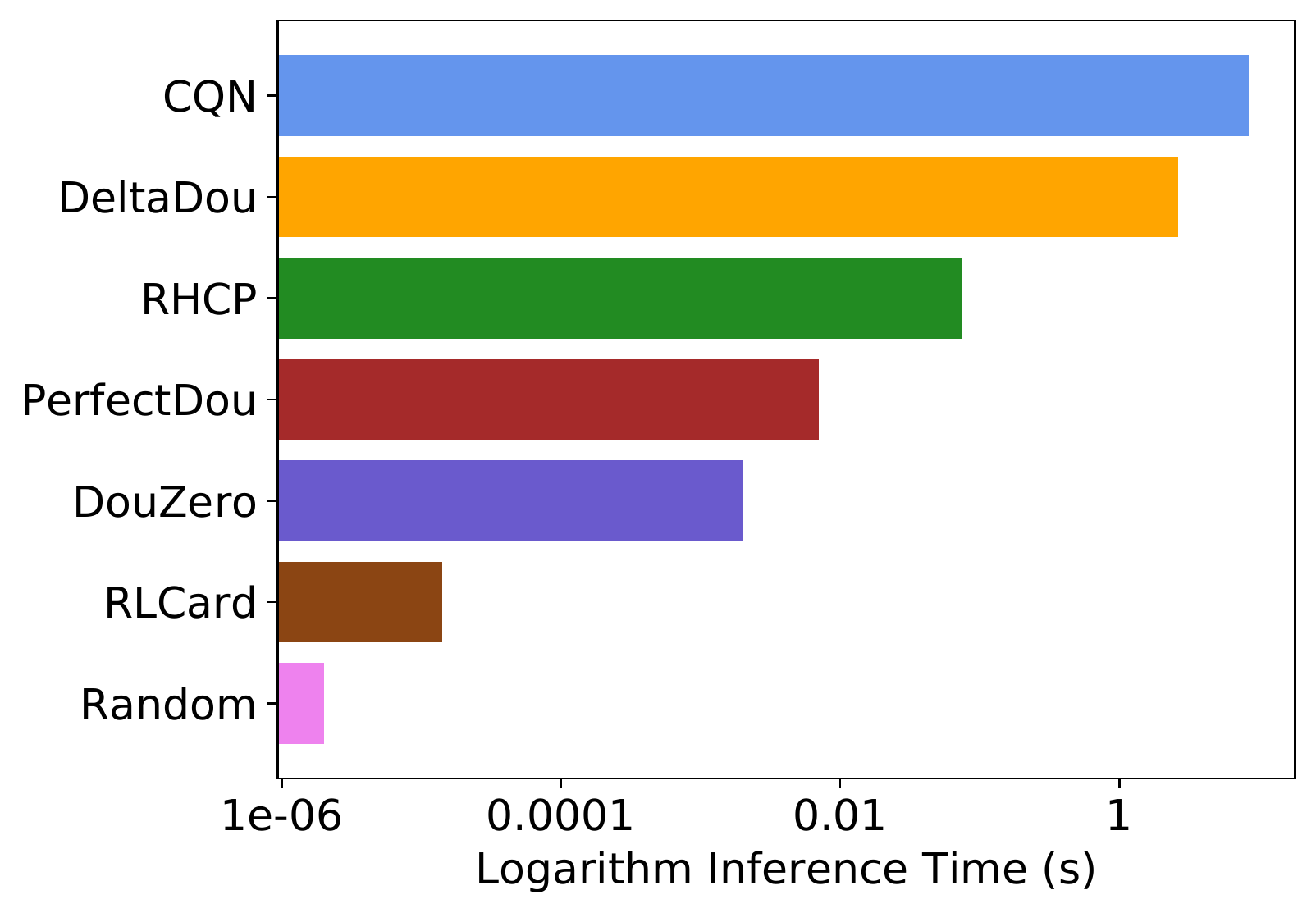}
\vspace{-20pt}
\caption{Comparison of the inference time.}
\vspace{-8pt}
\label{fig:infer}
\end{wrapfigure}
We further conduct runtime analysis to show the efficiency of PerfectDou w.r.t. the inference time (\textbf{RQ3}), which is reported in \fig{fig:infer}. All evaluations are conducted on a single core of Intel(R) Xeon(R) Gold 6130 CPU @ 2.10GHz. 
The inference time of each AI could be attributed to its pipeline and implementation in the playing time. CQN uses a large Q network (nearly 10$\times$ parameters larger than ours) with a complex card decomposer to derive reasonable hands. As a result, the inference time of CQN is the longest. Besides, both DeltaDou and RHCP-V2 contain lots of times of Monte Carlo simulations, thus slowing down the inference time. As comparisons, DouZero and PerfectDou only require one network forward inference time with a similar number of parameters. For RLCard, only handcraft rules are computed. Therefore, we can notice that PerfectDou is significantly faster than previous programs like DeltaDou, CQN and RHCP, yet is slightly slower than DouZero. To be more accurate, the average inference time of DouZero is 2 milliseconds compared with 6 milliseconds of PerfectDou. 
And the reason why PerfectDou is a bit slower than DouZero may due to the more complex feature processing procedure.
Note that this can be further optimized in practice such as changing the implementation from Python to C++, which is common for AI deployment. In addition, the model inference time of DouZero we test is 1.9 ms while PerfectDou is 4 ms, the difference is slight. The above analysis suggests that PerfectDou is applicable and affordable to real-world applications such as advanced game AI.

\subsection{In-depth Analysis and Case Studies}
In our experiments, we find that DouZero is leaky and unreasonable in many battle scenarios, while PerfectDou performs better therein. To quantitatively evaluate whether PerfectDou is stronger and more reasonable, we conduct an in-depth analysis by collecting the statistics among the games, and additionally illustrate some of the observations for qualitatively comparing the behavior of DouZero and PerfectDou. Detailed results can be further referred to \ap{sec:stats-analysis} and \ap{sec:case-study}, and here we list the key conclusions below:

The statistic analysis claims the rationality of PerfectDou:
(i) when playing as the \textit{Landlord}, PerfectDou plays fewer bombs to avoid losing scores and tends to control the game even when the \textit{Peasants} play more bombs; (ii) when playing as the \textit{Peasant}, two PerfectDou agents cooperate better with more bombs to reduce the control time of the \textit{Landlord} and its chance to play bombs; (iii) when playing as the \textit{Peasant}, the right-side \textit{Peasant} agent (play after the \textit{Landlord}) of PerfectDou throws more bombs to suppress the Landlord than DouZero, which is more like human strategy. From behavior observations, we also find: 1) DouZero is more aggressive but less thinking. 2) PerfectDou is better at guessing and suppressing. 3) PerfectDou is better at card combination. 4) PerfectDou is more calm. Beyond these, we also include battle results against skilled human players, additional training results and complete tournament results in the appendix.
\section{Conclusion} 

In this paper, we propose PerfectDou, a SoTA DouDizhu AI system that dominates the game. PerfectDou takes the advantage of the perfect-training-imperfection-execution (PTIE) training paradigm, and is trained within a distributed training framework. In experiments we extensively investigate how and why PefectDou can achieve the SoTA performance by beating all existing AI programs with reasonable strategic actions.
In fact, the PTIE paradigm is actually a variant of centralized-training-decentralized-execution (CTDE), applied for imperfect-information games in particular. Intuitively, PTIE is a general way for training imperfect-information game AI, with which we expect the value function can distill the perfect information to the policy which can only receive imperfect information. Although in this paper we only discuss its success on one of the hard poker games, DouDizhu, we believe it has the power to further improve the ability for other imperfect-information games, remaining a space of imagination to be explored by more future works.

\section*{Acknowledgement}
The SJTU team is partially supported by ``New Generation of AI 2030'' Major Project (2018AAA0100900) and National Natural Science Foundation of China (62076161).
Minghuan Liu is also supported by Wu Wen Jun Honorary Doctoral Scholarship, AI Institute, SJTU. We thank Ming Zhou for helpful discussions on game theory, and anonymous reviewers for constructive suggestions.

\bibliographystyle{plain}
\bibliography{ref}


\clearpage
\onecolumn
\newpage
\appendix
\section*{\huge Appendix}

\setcounter{theorem}{0}

\section{Additional Related Work}
\paragraph{Train with global information, test with local one.}
Utilizing global information to reduce the complexity of imperfect-information games has also been investigated in some works. For example, AlphaStar~\cite{vinyals2019grandmaster}, a grand-master level AI system for StarCraft II. In their implementation, the value network of the agent can observe the full information about the game state, including those that are hidden from the policy. They argue that such a training style improves training performance.
In our work, we formulate the idea as Perfect-Training-Imperfect-Execution (PTIE) or perfect information distillation technique for imperfect-information games, and show the effectiveness on complicated card games like DouDizhu.
Moreover, in Suphx~\cite{li2020suphx}, a strong Mahjong AI system, they used a similar method namely oracle guiding. Particularly, in the beginning of the training stage, all global information is utilized; then, as the training goes, the additional information would be dropped out slowly to none, and only the information that the agent is allowed to observe is reserved in the subsequent training stage. However, there are obvious difference between Suphx and PerfectDou. In Suphx, the perfect information is used by the actor and thus has to be dropped before the inference stage; on the contrary, PerfectDou feeds the critic with additional observations and distill the global information to the actor. 
Beyond games, Fang et al.~\cite{fang2021universal} worked on trading for order execution and proposed a different technique using global information other than PTIE, which trained a student policy with imperfect (real) market information and policy distillation from a teacher policy trained with perfect (oracle) market information.

\paragraph{Relation to sample-based CFR.} CFR aims to minimize the total regret of policy by minimizing the cumulative counterfactual regret in each infoset. The definition of regret highly relates to the definition of advantage used in RL community, which has been shown in lots of previous works \cite{srinivasan2018actor,fu2021actor}. Vanilla CFR~\cite{zinkevich2007regret} and many variants~\cite{waugh2015solving,brown2019deep} apply model-based approach to calculated all the weights of the game tree to update and obtain a good strategy (policy). However, when the game has long episodes and is hard for searching across the game tree, it is necessary to compute through trajectory samples, called sampled-based CFR methods~\cite{steinberger2020dream,gruslys2020advantage}. This resembles the learning procedure of RL algorithms. Recently, Fu et al.~\cite{fu2021actor} proposed a new form of sample-based CFR algorithm, and shown that PPO is exactly a practical implementation of it (but not PTIE), revealing close connections between CFR and RL. 

\section{More About DouDizhu}
\label{ap:doudizhu}
\subsection{Term of Categories}
\label{ap:terms}
In the work of Zha et al.~\cite{zhadouzero21}, they had shown a comprehensive introduction of DouDizhu game, so we think it may be wordy to repeat the stereotyped rules. However, for better understanding the cases shown in this paper, we introduce the typical term of categories in DouDizhu that are commonly used as follows. Note that all cards can suppress the cards in the same category with a higher rank, yet bomb can suppress any categories except the bomb with a higher rank. Rocket is the highest-rank bomb. \textbf{Kicker} refer to the unrelated or useless cards that players can deal out when playing some kind of categories of \textbf{main cards} (see below), which can be either a solo or a pair.
\begin{enumerate}
    \item \textbf{Solo}           : Any single card.
    \item \textbf{Pair}           : Two matching cards of equal rank.
    \item \textbf{Trio}           : Three individual cards of equal rank.
    \item \textbf{Trio with Solo} : Three individual cards of equal rank with a Solo as the kicker.
    \item \textbf{Trio with Pair} : Three individual cards of equal rank with a Pair as the kicker.
    \item \textbf{Chain of Solo}  : Five or more consecutive individual cards.
    \item \textbf{Chain of Pair}  : Three or more consecutive Pairs.
    \item \textbf{Chain of Trio (Plane)} : Two or more consecutive Trios.
    \item \textbf{Plane with Solo}: Two or more consecutive Trios with each has a distinct individual kicker card and Plane as the main cards.
    \item \textbf{Quad with Solo} : Four-of-a-kind with two Solos as the kicker and Four-of-a-kind as the main cards.
    \item \textbf{Quad with Pair} : Four-of-a-kind with a set of Pair as the kicker and Quad with Pair as the main cards.
    \item \textbf{Quad with Pairs} : Four-of-a-kind with two sets of Pair as the kicker and Quad with Pairs as the main cards.
    \item \textbf{Bomb}           : Four-of-a-kind.
    \item \textbf{Rocket}         : Red and black jokers.
\end{enumerate}

\subsection{Scoring Rules}
\label{ap:score}
In Zha et al.~\cite{zhadouzero21}, they pay more attention to the win/lose result of the game but care less about the score. However, in real competitions, players must play for numbers of games and are ranked by the scores they win. And that is why we think ADP is a better metric for evaluating DouDizhu AI systems because a bad AI player can win a game with few scores but lose with much more scores.

Specifically, in each game, the \textit{Landlord} and the \textit{Peasants} have base scores of 2 and 1 respectively. When there is a bomb shown in a game, the score of each player doubles. For example, a \textit{Peasant} player first shows a bomb of $4$ and then the \textit{Landlord} player suppresses it with a rocket, then the base score of each \textit{Peasant} becomes 4 and the \textit{Landlord} becomes 8. A player will win all his scores after winning the game, or loses all of them vice versa.


\section{Additional System Design Details}

\subsection{Card Representation Details}
\label{ap:card-representation}

In the system of PerfectDou, we augment the basic card in hand matrix with explicitly encoded card types as additional features, in order to allow the agent realizing the different properties of different kind of cards. The size details are shown in \tb{tb:card-representation}.

\begin{table}[h]
\begin{center}
\caption{Card representation design.}
\label{tb:card-representation}
\begin{small}
\begin{sc}
\resizebox{0.3\columnwidth}{!}{
\begin{tabular}{cc}
\toprule
        \textbf{Card Matrix Feature} & \textbf{Size} \\
\midrule
  card in hand & $4  \times 15$ \\
  solo & $1  \times 15$ \\
  pair & $1  \times 15$ \\
  trio & $1  \times 15$ \\
  bomb & $1  \times 15$ \\
  rocket & $1  \times 15$ \\ 
  chain of solo  & $1 \times 15$ \\
  chain of pair & $1 \times 15$ \\
  chain of trio & $1 \times 15$ \\
\bottomrule
\end{tabular}
}
\end{sc}
\end{small}
\end{center}
\end{table}

\subsection{Action Feature Details}

\begin{table}[h]
\begin{center}
\caption{Action feature design.}
\label{tb:action-feature}
\begin{sc}
\resizebox{0.7\columnwidth}{!}{
\begin{tabular}{cc}
\toprule
  \textbf{Feature Design} & \textbf{Size} \\
\midrule
  card matrix of action & $12 \times 15$ \\
  if this action is bomb & $1$ \\
  if this action is the largest one & $1$ \\
  if this action equals the number of left player's cards in hand  & $1$ \\
  if this action equals the number of right player's cards in hand  & $1$ \\
  the minimum steps to play-out all left cards after this action played  & $1$ \\
  if this action is valid & $1$ \\
  action id & $1$ \\
\bottomrule
\end{tabular}
}
\end{sc}
\end{center}
\end{table}

The action features are a flatten matrix from $12 \times 15$ action card matrix plus $1 \times 7$ extra dimensions describing the property of the cards as shown in \tb{tb:action-feature}. Since the number of actions in each game state varies, which can lead to different lengths of action features, a fixed length matrix is flattened to store all action features where the non-available ones are marked as zero.

\subsection{Value Network Structure}
\label{ap:value-structure}
\begin{figure}[htbp]
\centering
\vspace{-5pt}
\includegraphics[width=0.7\columnwidth]{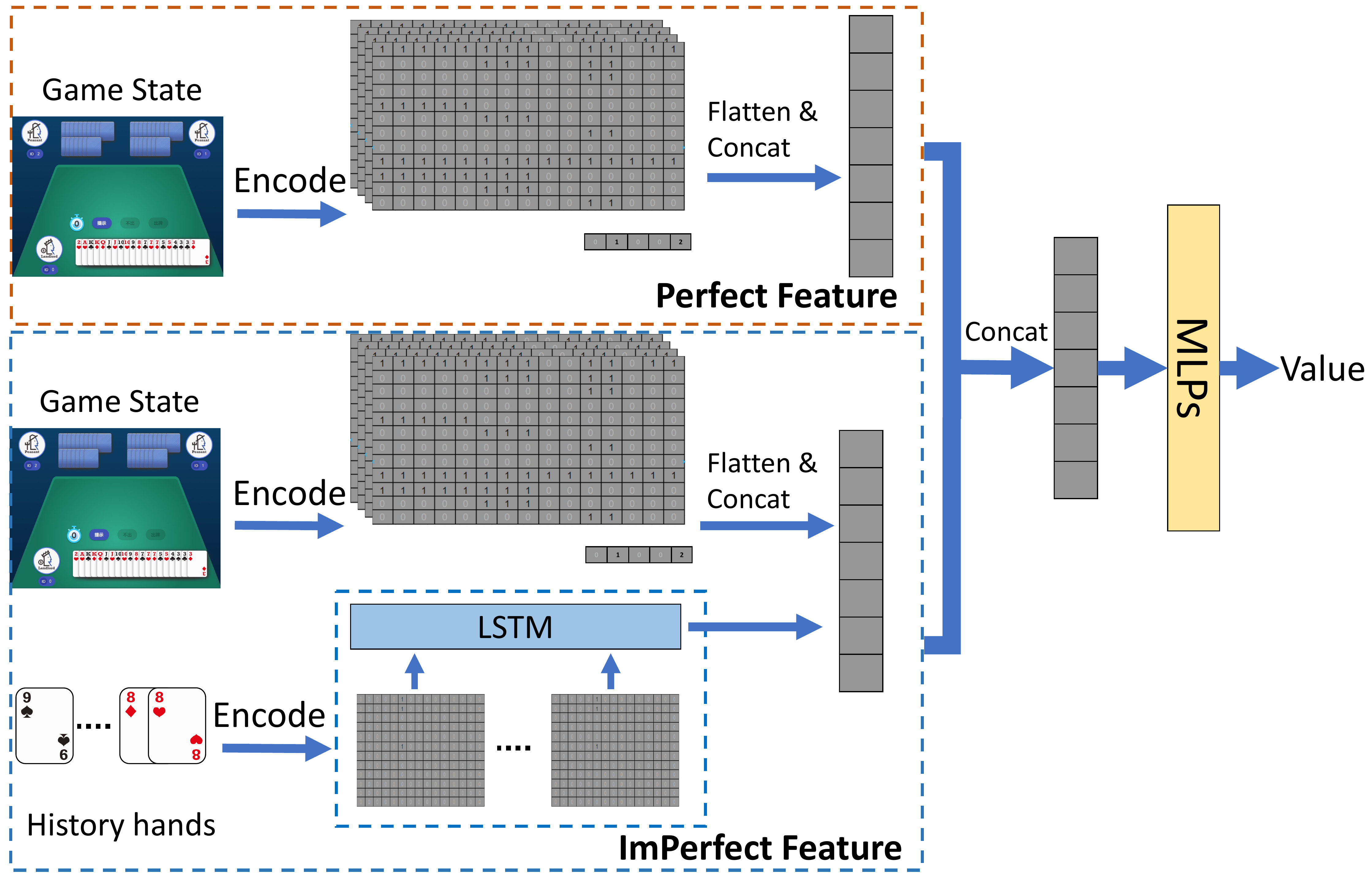}
\vspace{-3pt}
\caption{The value network structure of the proposed PerfectDou system. The network predicts values using both the imperfect feature and the perfect feature and distill the knowledge into the policy in the training.}
\label{fig:value-structure}
\end{figure}

The value network of PerfectDou is designed to evaluate the current situation of players, and we expect that the value function can utilize the global information, in other words, know the exact node the player is in. Therefore, we should feed additional information that the policy is not allowed to see in our design. Specifically, as shown in \fig{fig:value-structure}, the imperfect feature for indistinguishable nodes is encoded using the shared network as in the policy network; besides, we also encode the perfect feature of distinguishable nodes that the policy cannot observe during its game playing. The encoded vector are then concatenated to a simple MLP to get the scalar value output.

\section{Experiments}
\label{ap:exps}

\subsection{Setups, Hyperparameters and Training Details}
In our implementation, a small distributed training cluster is built using 880 CPUs cores and 8 GPUs. Horovod~\cite{sergeev2018horovod} is used to synchronize gradients between GPUs, the total batch size is 1024, 128 for each GPU. The most important hyperparameters in our experiment are shown in \tb{tab:hyper}. Specifically, in our design, we simplify the discrete action space from 27472 (include all possible combinations) into an abstract action space of 621 for learning the actor, followed by an decoding strategy to get the final action (see \ap{ap:action-space} for more details). 

During self-play training, we find a better practical solution for DouDizhu is to keep three different models for \textit{Landlord} and two \textit{Peasants} separately which is only updated by their own data against the latest opponent model. In the main training stage, the total reward function will be a basic reward (in this paper we use ADP all the time) augmented with the designed oracle reward as shown in \se{sec:reward}, which is found to be extremely useful for accelerating convergence. In the final stage, the oracle reward is removed and only the ADP reward is used to fine-tune the model for reaching a better performance measured by the ADP metric.

\begin{table}[htbp]
  \centering
  \caption{Hyperparameters. $^*$ refers to the maximum version gap allowed between the models used for sampling and training.}
  \label{tab:hyper}%
    \begin{tabular}{|l|c|}
    \hline
    Learning rate & 3e-4\\
    \hline
    Optimizer & {Adam} \\
    \hline
    Discount factor $\gamma$ & {1.0} \\
    \hline
    $\lambda$ of GAE & 0.95 \\
    \hline
    Step of GAE & 24 (8 for each player) \\
    \hline
    Batch size & {1024} \\
    \hline
    Entropy weight of PPO & 0.1 \\
    \hline
    Length of LSTM & 15 (5 for each player) \\
    \hline
    Max model lag$^*$ & 1 \\
    \hline
    Intermediate reward scale & 50 \\
    \hline
    Policy MLP hidden sizes & [256, 256, 256, 512] \\
    \hline
    Policy MLP output size (action space size) & 621 \\
    \hline
    Value MLP hidden sizes & [256, 256, 256, 256] \\
    \hline
    Value MLP output size & 1 \\
    \hline
    \end{tabular}%
\end{table}%

\begin{table*}[hbtp]
    \centering
    \caption{Average per game statistics of important behaviors over 100k decks: \texttt{Game Len} is the average number of rounds in a game; \texttt{\% Bomb} represents the average percentage of bombs (a type of card can suppress any categories except the bomb with a higher rank, see \protect\ap{ap:doudizhu}) played in the game; \texttt{Left} and \texttt{Right} are the relative position to the \textit{Landlord}; and \texttt{Landlord Control Time} measures the number of rounds that the landlord plays an action suppressing all other players.}
    \label{tab:statistics}
    \vspace{-0pt}
    \resizebox{0.99\textwidth}{7mm}{
    \begin{tabular}{c|ccccccc|c}
    \toprule
    \textit{Landlord} Agent & WP & ADP & \texttt{Game Len} & \texttt{\%Bomb of Left} \textit{Peasant} & \texttt{\%Bomb} of \textit{Landlord} & \texttt{\%Bomb of Right} \textit{Peasant} & \texttt{Landlord Control Time} & \textit{Peasant} Agent\\
    \hline
    \midrule
     PerfectDou (2.5e9) & 0.446 & -0.407 & 33.347 & 68.05 & 28.46 & 74.90 & 12.993 & \multirow{2}{*}{DouZero ($\sim$1e10)}\\
     DouZero ($\sim$1e10) & 0.421 & -0.461 & 33.911 & 66.24 & 28.73 & 75.29 & 9.005 & \\
     \midrule
     PerfectDou (2.5e9) & 0.387 & -0.608 & 31.157 & 66.13 & 26.67 & 79.68 & 10.518 & \multirow{2}{*}{PerfectDou (2.5e9)}\\
     DouZero ($\sim$1e10) & 0.360 & -0.686 & 31.267 & 64.80 & 26.72 & 79.29 & 7.123 & \\
     \bottomrule
    \end{tabular}
    }
    \vspace{-9pt}
\end{table*}

\subsection{In-Depth Statistical Analysis}
\label{sec:stats-analysis}
In our experiments, we find that DouZero is leaky and unreasonable in many battle scenarios, while PerfectDou performs better therein. To quantitatively evaluate whether PerfectDou is stronger and more reasonable, we conduct an in-depth analysis and collect the statistics among the games between DouZero and PerfectDou. Particularly, we organize games between PerfectDou and Douzero to play in different roles for 100,000 decks in each setting. Since the roles are assigned randomly instead of opting by agents themselves in our experiments, and the \textit{Landlord} has a higher base score with three extra cards, we observe that playing as a \textit{Landlord} is always harder to win and leads to negative ADPs. From the statistics shown in \tb{tab:statistics}, we learn many lessons about the rationality of PerfectDou: 
(i) when playing as the \textit{Landlord}, PerfectDou plays fewer bombs to avoid losing scores and tends to control the game even the \textit{Peasants} play more bombs; (ii) when playing as the \textit{Peasant}, two PerfectDou agents cooperate better with more bombs to reduce the control time of the \textit{Landlord} and its chance to play bombs; (iii) when playing as the \textit{Peasant}, the right-side \textit{Peasant} agent (play after the \textit{Landlord}) of PerfectDou throws more bombs to suppress the Landlord than DouZero, which is more like human strategy.

\subsection{Case Study: Behavior of DouZero vs PerfectDou}
\label{sec:case-study}
In this section, we list some of the observations during the games for comparing the behavior of DouZero and PerfectDou to qualitatively support our analysis.

\begin{figure*}[!t]
\centering
\subfigure[Case study: DouZero is more aggressive by choosing a chain of solo in the beginning but leaves the pair of 3 in the hand.]{
\begin{minipage}[b]{0.4\linewidth} 
\label{fig:case1-1}
\includegraphics[width=1.1\columnwidth]{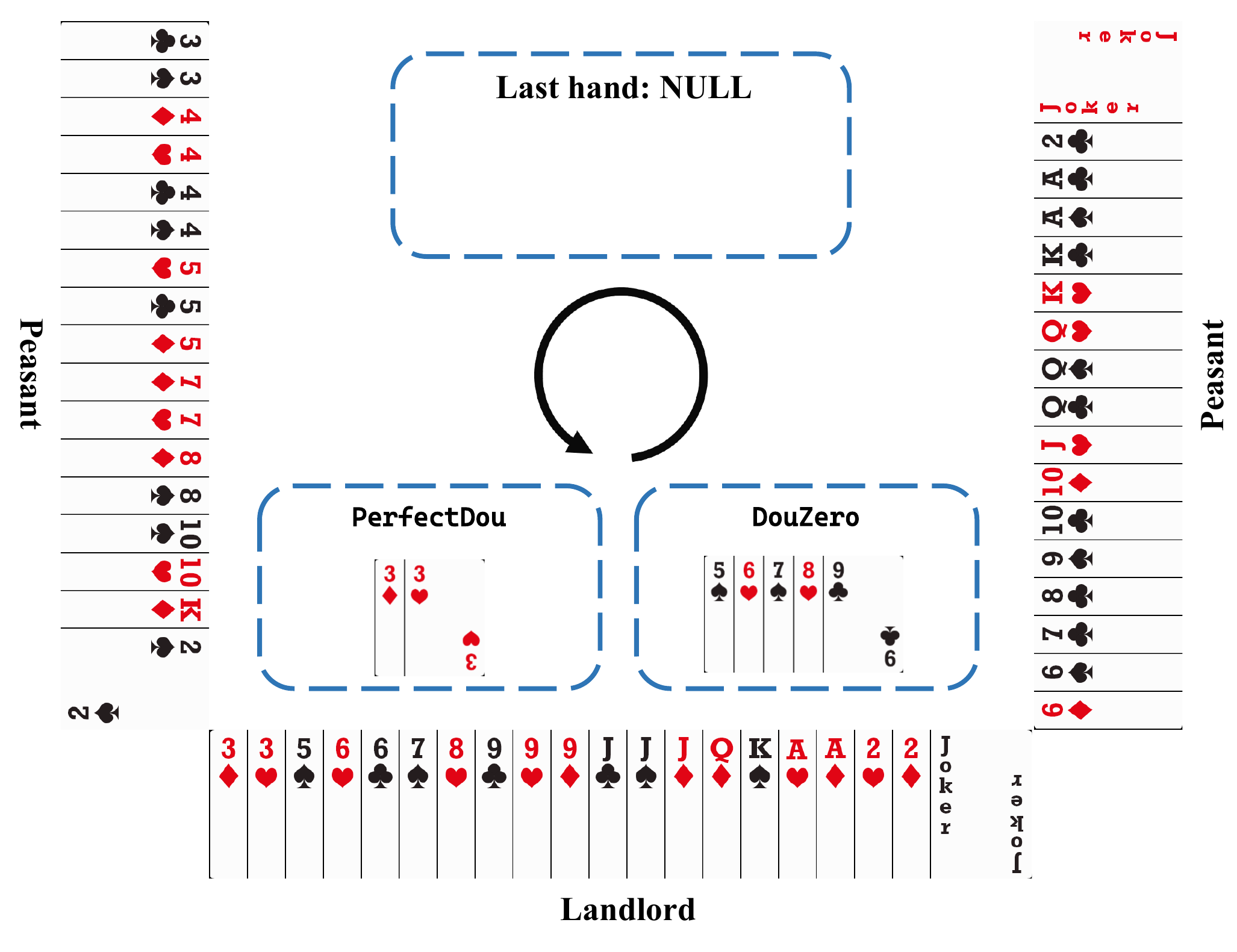}
\end{minipage}
}
\hspace{7pt}
\subfigure[Case study: DouZero is more aggressive by suppressing the \textit{Landlord} but less thinking on the consequence of the left hands of solos.]{
\label{fig:case1-2}
\begin{minipage}[b]{0.4\linewidth} 
\includegraphics[width=1.1\columnwidth]{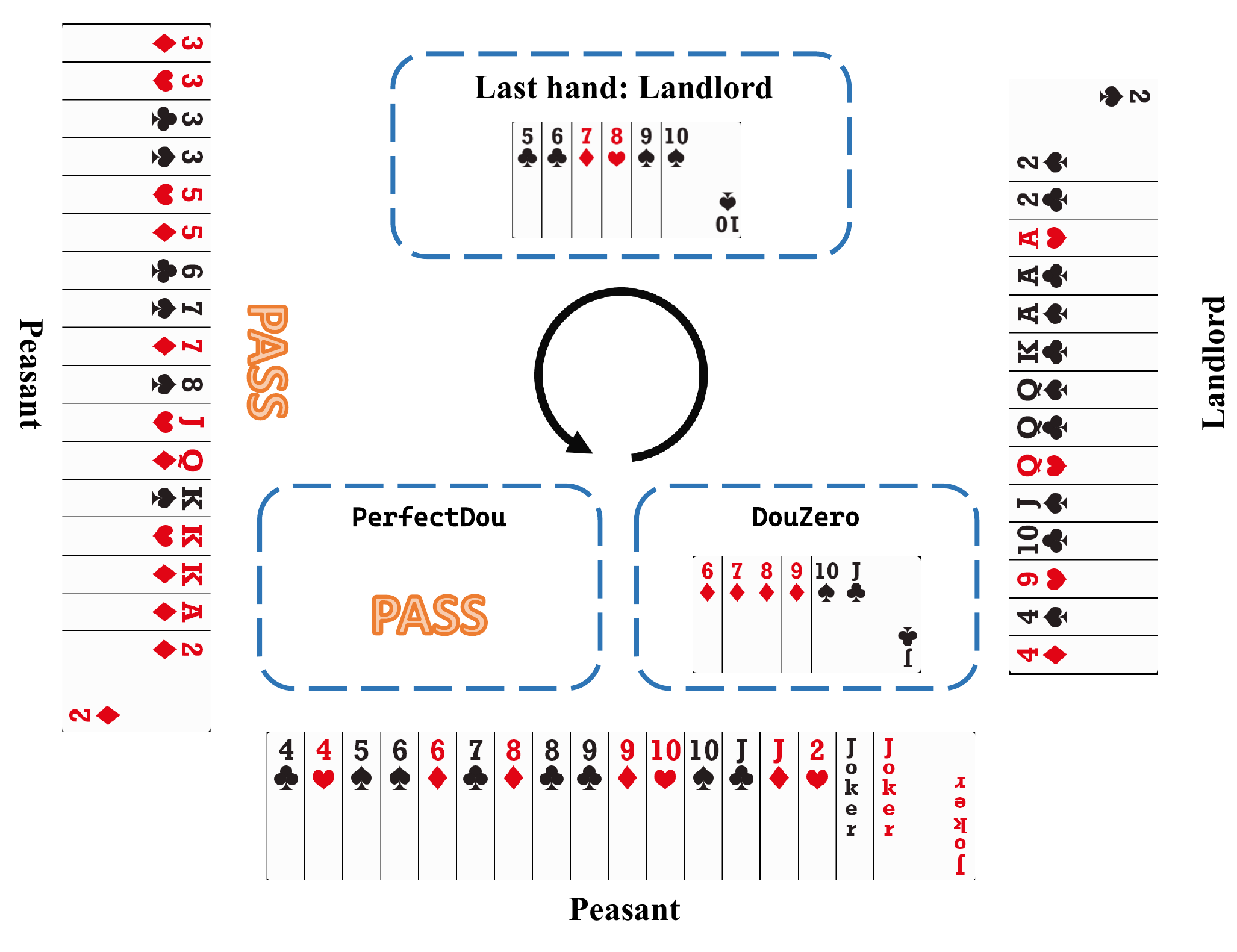}
\end{minipage}
}
\subfigure[Case study: the teammate shows a pair of T and DouZero chooses to pass; on the contrary, PerfectDou chooses suppressing by a pair of Q -- the minimal pair of the opponent.]{
\begin{minipage}[b]{0.4\linewidth} 
\label{fig:case4-1}
\includegraphics[width=1.1\columnwidth]{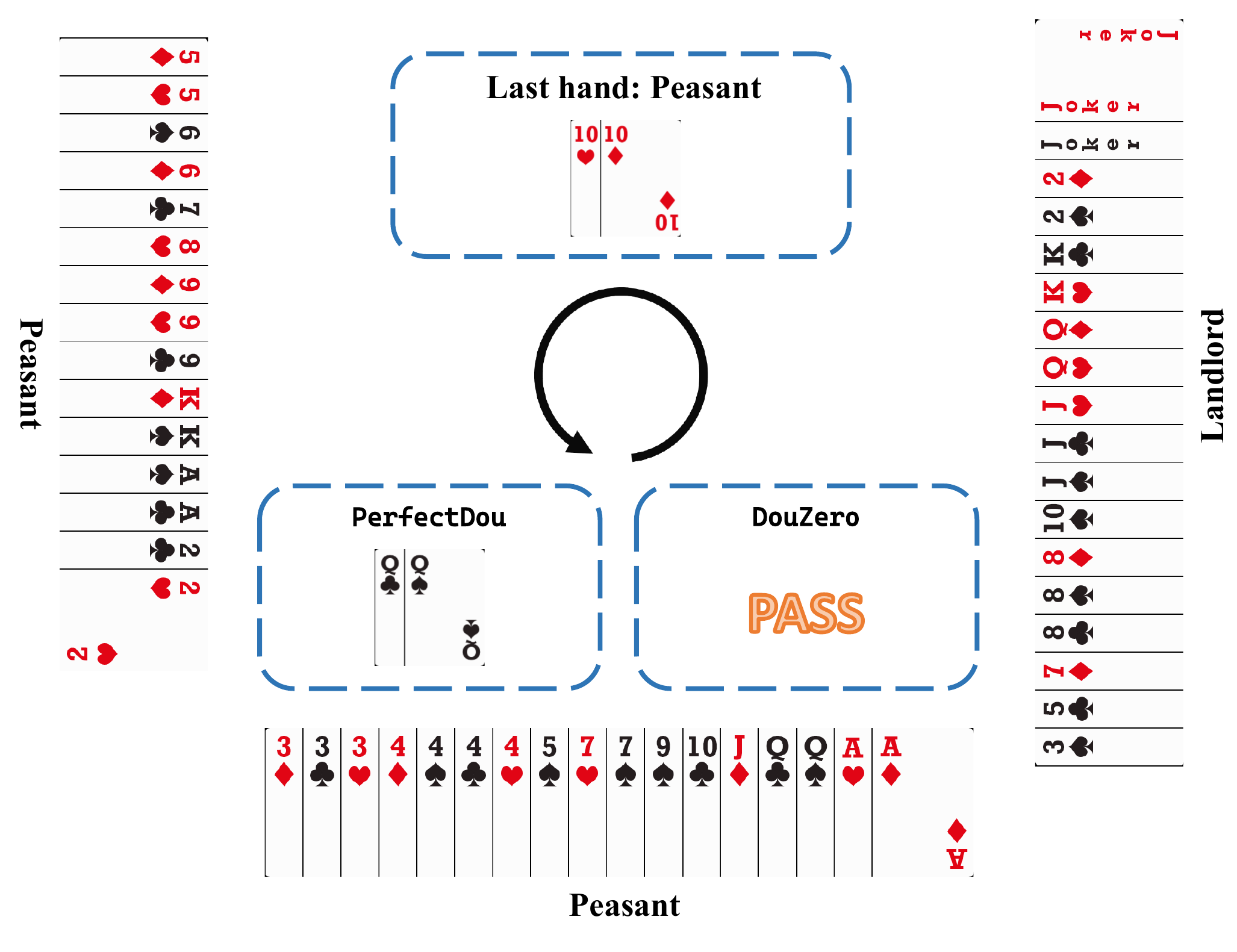}
\end{minipage}
}
\hspace{7pt}
\subfigure[Case study: PerfectDou chooses to split the plane (999, $TTT$) since it considers there is a chain of solo ($9TJQK$) left.]{
\begin{minipage}[b]{0.4\linewidth} 
\label{fig:case2-1}
\includegraphics[width=1.1\columnwidth]{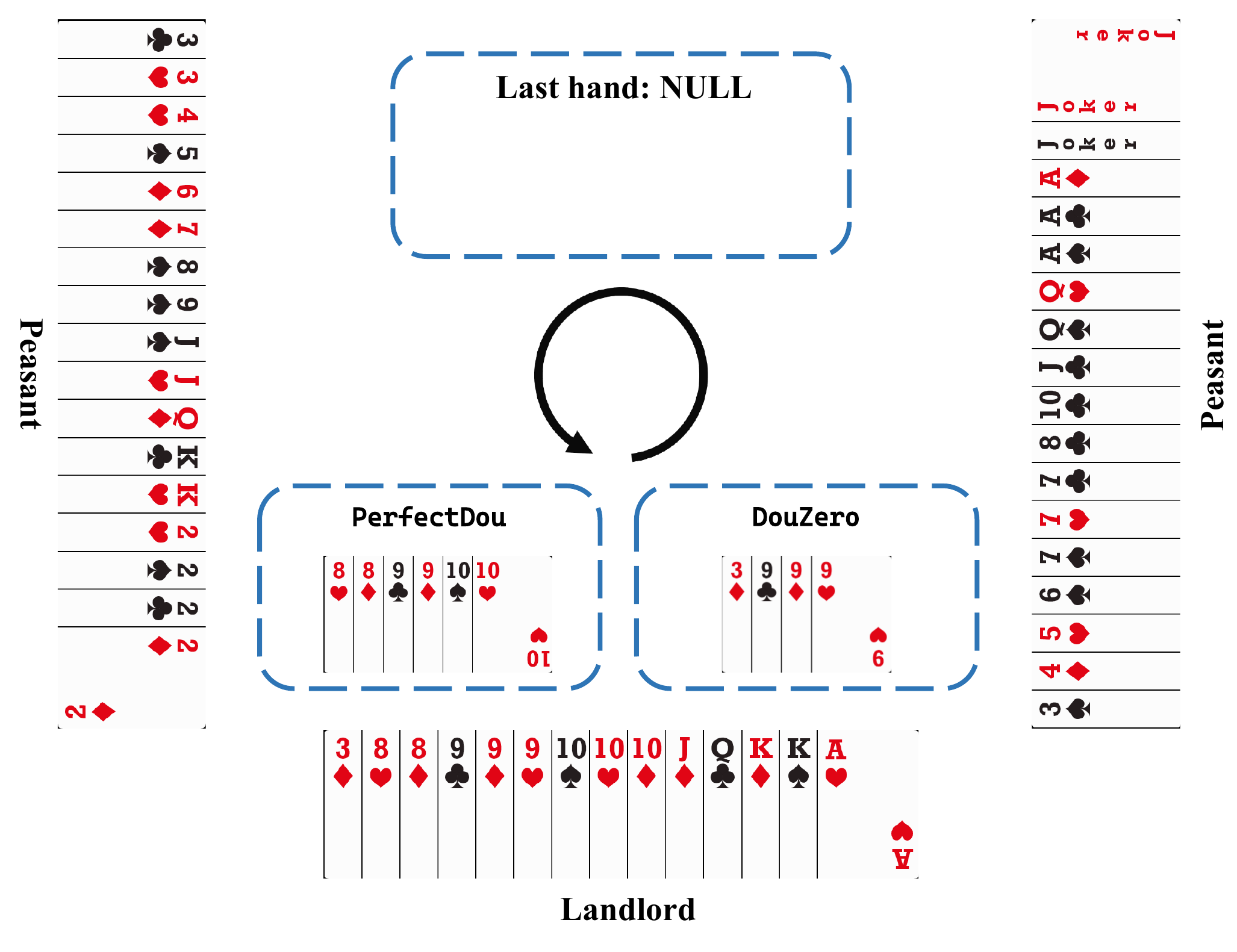}
\end{minipage}
}
\subfigure[Case study: DouZero splits the rocket bomb while PerfectDou chooses to keep it.]{
\begin{minipage}[b]{0.48\linewidth} 
\label{fig:case3-1}
\includegraphics[width=1\columnwidth]{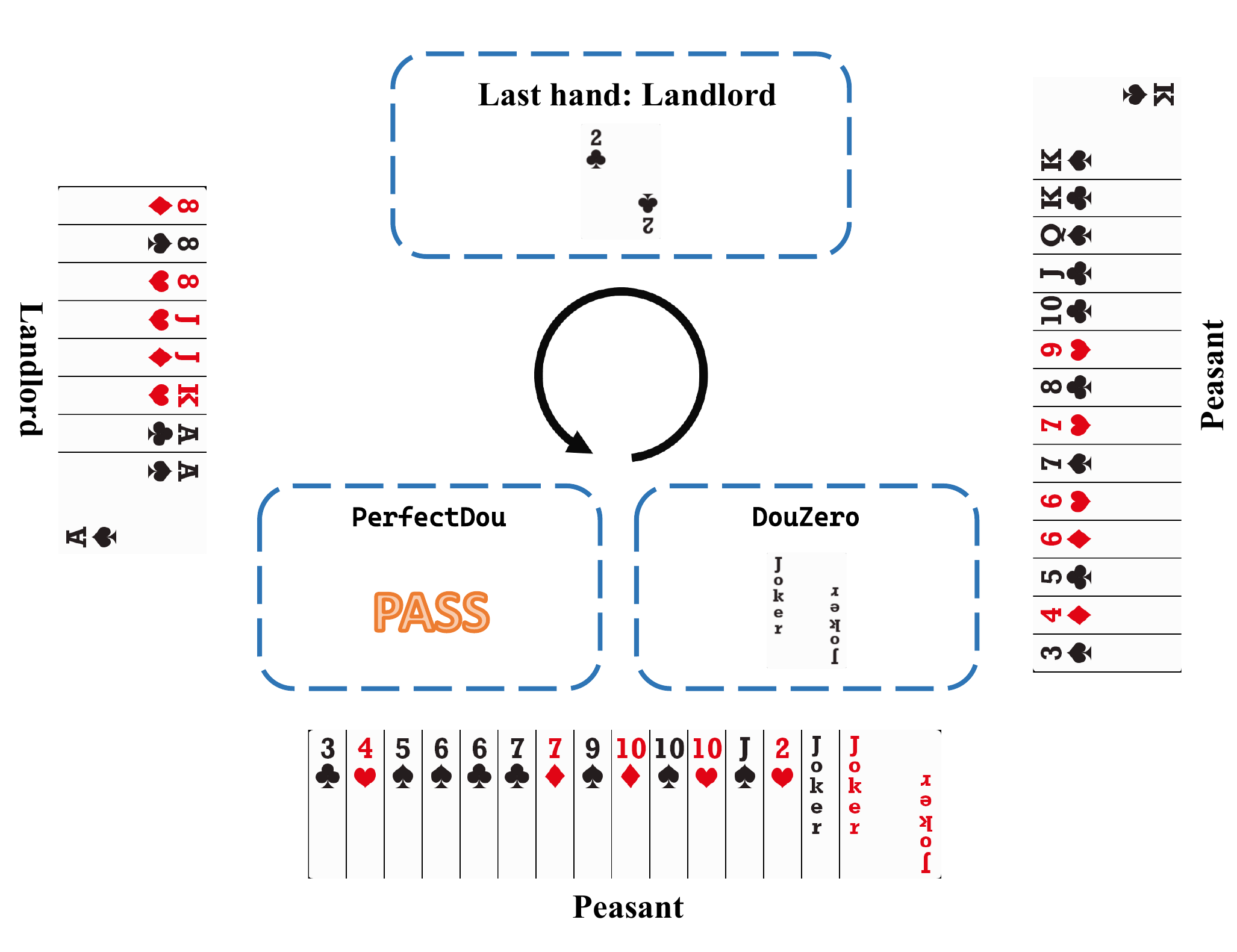}
\end{minipage}
}
\vspace{-3pt}
\caption{Case studies.}
\vspace{-2pt}
\label{fig:heat_maps}
\end{figure*}

\paragraph{DouZero is more aggressive but less thinking.} 
The first observation is that DouZero is extremely aggressive without considering the left hands. For instance, as shown in \fig{fig:case1-1}, in the beginning DouZero chooses a chain of solo but leaves the pair of 3, which can be dangerous since the pair of 3 is the one of the minimum cards and cannot suppress any card; \fig{fig:case1-2} illustrates another strong case, where DouZero also chooses a chain of solo to suppress the opponent without considering the consequence of leaving a hand of solos. On the contrary, PerfectDou is more conservative and steady. We believe the proposed perfect information distillation mechanism helps PerfectDou to infer global information in a more reasonable way.

\paragraph{PerfectDou is better at guessing and suppressing.} 
We observe another fact that the usage of perfect information distillation within the PTIE framework benefits PerfectDou a lot by suppressing the opponents in advance. In \fig{fig:case4-1} shows a case when the teammate puts a pair of $T$\footnote{We denote $T$(en) as the card $10$ for simplicity.}, DouZero chooses to pass; on the contrary, PerfectDou chooses suppressing by a pair of $Q$ -- the minimal pair of the \textit{Landlord}.


\paragraph{PerfectDou is better at card combination.} 
In the battle shown in \fig{fig:case2-1}, PerfectDou shows the better ability on the strategy of card combination. Specifically, PerfectDou chooses to split the plane (999, $TTT$ since it considers there is a chain of solo  ($9TJQK$) left. However, DouZero only takes the trio, which will be easily suppressed by the opponent. This benefits from the proper design of the card representation and the action feature of PerfectDou.


\paragraph{PerfectDou is more calm.} 

\fig{fig:case3-1} depicts a typical and interesting scenario where PerfectDou shows its calm and careful consideration over the whole. In the game, the last hand is of the \textit{Landlord} with a solo 2, and it only has 8 cards left in the hand. DouZero seems afraid and splits the rocket bomb; however, PerfectDou benefits from the advantage reward design and is calm considering there is a greater chance on winning the game with a higher score by keeping the bomb. 


\subsection{Battle Results Against Skilled Human Players}
We further invite some skilled human players to play against PerfectDou. Particularly, each human player plays with two AI players. In other words, each game is involved with either two AI \textit{Peasants} against one human \textit{Landlord}, or one AI \textit{Peasant} cooperating with one human \textit{Peasant} against one AI \textit{Landlord}. The results are shown in \tb{tb:human}. One can easily observe that PerfectDou takes evident advantage during the game.

\begin{table}[htbp]
    \centering
    \caption{Battle results against skilled human players for 1260 episodes of game.}
    \label{tb:human}
    \small
    \resizebox{0.4\columnwidth}{8mm}{
    \begin{tabular}{l|cc}
    \toprule
    \multirow{2}{*}{\diagbox [width=9em,trim=l] {\texttt{A}}{\texttt{B}}} & \multicolumn{2}{c}{Skilled human}\\
    \cline{2-3}
    & WP & ADP \\
    \hline
    \midrule
     PerfectDou (2.5e9) & 0.625 & 0.590 \\
     \bottomrule
    \end{tabular}
    }
    \vspace{-9pt}
\end{table}

\subsection{Additional Training Results}
\label{ap:curves} 
\begin{figure}[htbp]
\centering
\vspace{-0pt}
\includegraphics[width=0.4\columnwidth]{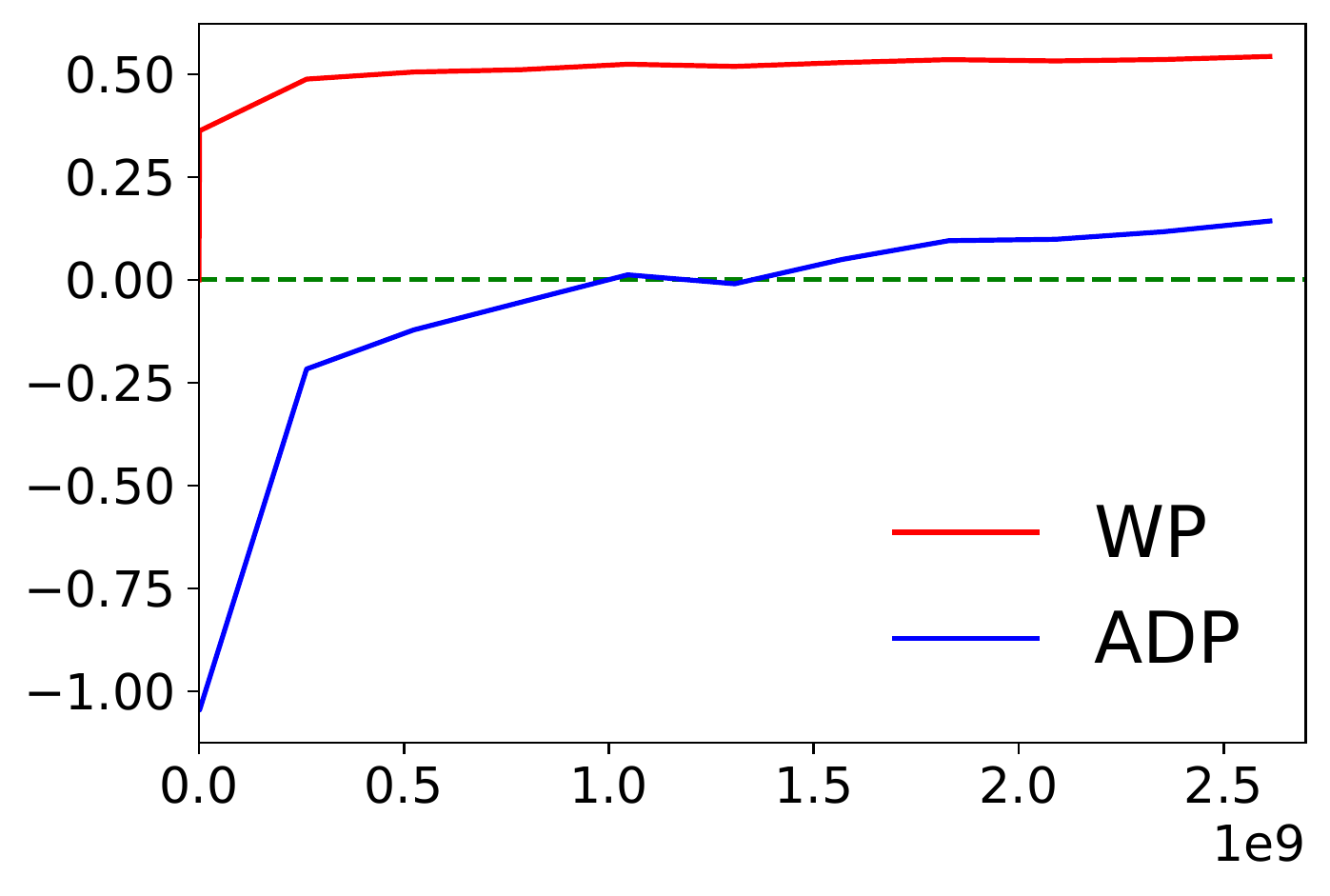}
\vspace{-10pt}
\caption{Learning curves of WP and ADP against the final model of DouZero w.r.t. timesteps for PerfectDou. Every evaluation contains 10,000 decks. PerfectDou is able to beat DouZero without considering the scores at the beginning of the training, around $1.5e6$ steps.}
\label{fig:wp-adp-curve}
\end{figure}

\fig{fig:wp-adp-curve} shows the learning curves of WP and ADP against DouZero for PerfectDou with a single run, and every evaluation contains 10,000 decks. As shown in the figure, PerfectDou can easily beat DouZero (on WP) without considering the scores (ADP) at the beginning of the training; but after 1.5e9 steps of training, PerfectDou is able to fully beat DouZero (both WP and ADP).

\subsection{Extended Competition against DouZero}
\paragraph{Cycling Different Hands for the \textit{Landlord}.}
In our main competition conducted in \se{sec:comp_eval}, all games are randomly generated and played twice with the same assigned hand cards for the \textit{Landlord}, once each algorithm controlling the \textit{Landlord} and once two \textit{Peasants}. 
In this section, we test cycling the 3 hand cards of every randomly generated game for the \textit{Landlord}, and test PerfectDou against DouZero by controlling the \textit{Landlord} separately, which leads to 6 times of battle. In this situation, we obtain the results as follows showing that PerfectDou has consistent advantage (\tb{tb:cycling}).

\begin{table}[htbp]
    \centering
    \caption{Results of cycling different hands for Landlord by playing 100k decks.}
    \label{tb:cycling}
    \small
    \resizebox{0.6\columnwidth}{10mm}{
    \begin{tabular}{l|cccc}
    \toprule
    \multirow{2}{*}{\diagbox [width=9em,trim=l] {\texttt{A}}{\texttt{B}}} & \multicolumn{2}{c}{PerfectDou} & \multicolumn{2}{c}{DouZero}\\
    \cline{2-5}
    & WP & ADP \\
    \hline
    \midrule
     PerfectDou & - & - & 0.544 & 0.150 \\
     DouZero & 0.456 & -0.150 & - & - \\
     \bottomrule
    \end{tabular}
    }
    \vspace{-9pt}
\end{table}

\paragraph{\textit{Peasants} Paired with Different AIs.} We also include an interesting battle by pairing the peasants with different algorithms. Specifically, DouZero or PerfectDou plays the \textit{Landlord} while one \textit{Peasant} is assigned with DouZero and the other \textit{Peasant} is played by PerfectDou. The results are concluded in \tb{tb:paired}, which reveal that when playing as the peasant, PerfectDou can better cooperate with its teammate than DouZero. And when playing as the landlord, PerfectDou outperforms DouZero against all types of opponents.

\begin{table}[htbp]
    \centering
    \caption{Results of battles that the \textit{Peasants} are paired with different AIs by 100k decks and 6 times of battle per deck. The results are evaluated from the \textit{Landlord} side.}
    \label{tb:paired}
    \small
    \resizebox{0.65\columnwidth}{8mm}{
    \begin{tabular}{l|cccccc}
    \toprule
    \multirow{2}{*}{\diagbox [width=9em,trim=l] {Landlord}{Peasant}} & \multicolumn{2}{c}{PerfectDou + DouZero} & \multicolumn{2}{c}{PerfectDou} &  \multicolumn{2}{c}{DouZero}\\
    \cline{2-7}
    & WP & ADP & WP & ADP & WP & ADP \\
    \hline
    \midrule
     PerfectDou & 0.424 & -0.448 & 0.389 & -0.606 & 0.452 & -0.375

 \\
     DouZero & 0.395 & -0.534 & 0.363 & -0.676 & 0.421 & -0.465

 \\
     \bottomrule
    \end{tabular}
    }
    \vspace{-9pt}
\end{table}

\subsection{Complete Tournament Results of ADP for \textit{Landlord} and \textit{Peasants}}

We report the complete tournament results of ADP and WP for \textit{Landlord} and \textit{Peasants} in \tb{tab:complete-adp} and \tb{tab:complete-wp}. PerfectDou tends to have more advantage of \textit{Peasants} than that of \textit{Landlord}, especially when competes against stronger baselines. We believe that the proposed perfect information distillation technique allows for better cooperation between two \textit{Peasants}. In addition, since the roles are assigned instead of opting according to hand in our competition, and the \textit{Landlord} has extra three cards and can lose a higher base score, the \textit{Peasants} seems having more chance to win the game. Therefore, almost all methods can play better results as a \textit{Peasant} than that as a \textit{Landlord}.

\begin{table*}[htb]
    \centering
    \caption{ADP results of DouDizhu tournaments for existing AI programs by playing 10k decks. L: ADP of \texttt{A} as Landlord; P: ADP of \texttt{A} as Peasants. Algorithm \texttt{A} outperforms \texttt{B} if the ADP of L or P is larger than 0 (highlighted in boldface). 
    We note that DouZero is the current SoTA DouDizhu bot. Numerical results except marked $*$ are directly borrowed from Zha et al.~\protect\cite{zhadouzero21}.}
    \label{tab:complete-adp}
    \vspace{-4pt}
    \small
    \vspace{-0pt}
    \setlength{\tabcolsep}{2.0pt}
    \resizebox{0.99\textwidth}{14mm}{
    \begin{tabular}{c||l|cc|cc|cc|cc|cc|cc}
    \toprule
     
\multirow{2}{*}{Rank} & \multirow{2}{*}{\diagbox [width=8em,trim=l] {\texttt{A}}{\texttt{B}}} & \multicolumn{2}{c|}{PerfectDou}& \multicolumn{2}{c|}{DouZero} & \multicolumn{2}{c|}{DeltaDou} & \multicolumn{2}{c|}{RHCP-v2} & \multicolumn{2}{c|}{CQN} &\multicolumn{2}{c}{Random}\\
\cline{3-14}
& & P & L & P & L & P & L & P & L & P & L & P & L \\
    \hline
    \midrule
     1 & PerfectDou (Ours) & 0.656$^*$ & -0.656$^*$ & \textbf{0.686}$^*$ & -0.407$^*$ & \textbf{0.980}$^*$ & -0.145$^*$ & \textbf{0.872}$^*$ & \textbf{0.138}$^*$ & \textbf{2.020}$^*$ & \textbf{2.160}$^*$ & \textbf{3.008}$^*$ & \textbf{3.283}$^*$\\
     2 & DouZero (Public) & 0.407$^*$ & -0.686$^*$ & 0.435$^*$ & -0.435$^*$ & \textbf{0.858}$^*$ & -0.342$^*$ & \textbf{0.166}$^*$ & -0.046$^*$ & \textbf{2.001}$^*$ & \textbf{1.368}$^*$ & \textbf{2.818}$^*$ & \textbf{3.254}$^*$\\
     3 & DeltaDou & \textbf{0.145}$^*$ & -0.980$^*$ & \textbf{0.342}$^*$ & -0.858$^*$ & \textbf{0.476} & -0.476 & \textbf{1.878}$^*$  & \textbf{0.974}$^*$  & \textbf{1.849} & \textbf{1.218} & \textbf{2.930} & \textbf{3.268}\\
     4 & RHCP-v2 & -0.138$^*$ & -0.872$^*$ & \textbf{0.046}$^*$ & -0.166$^*$ & -0.974$^*$ & -1.878$^*$  & \textbf{0.182}$^*$ & -0.182$^*$ & \textbf{1.069}$^*$ & \textbf{1.758}$^*$ & \textbf{2.560}$^*$ & \textbf{2.780}$^*$\\
     5 & CQN & -2.160$^*$ & -2.020$^*$ & -1.368$^*$ & -2.001$^*$ & -1.218 & -1.849 & -1.758$^*$ & -1.069$^*$ & \textbf{0.056} & -0.056 & \textbf{1.992} & \textbf{1.832}\\
     6 & Random & -3.283$^*$ & -3.008$^*$ & -3.254$^*$ & -2.818$^*$ & -3.268 & -2.930  & -2.780 & -2.560$^*$ & -1.832 & -1.991 & \textbf{0.883} & -0.883 \\
     \bottomrule
    \end{tabular}
    }
\end{table*}

\begin{table*}[htb]
    \centering
    \caption{WP results of DouDizhu tournaments for existing AI programs by playing 10k decks. L: WP of \texttt{A} as Landlord; P: WP of \texttt{A} as Peasants. Algorithm \texttt{A} outperforms \texttt{B} if the WP of L or P is larger than 0.5 (highlighted in boldface). Numerical results except marked $*$ are directly borrowed from Zha et al.~\protect\cite{zhadouzero21}.}
    \vspace{-5pt}
    \small
    \vspace{-0pt}
    \setlength{\tabcolsep}{2.0pt}
    \resizebox{0.99\textwidth}{14mm}{
    \begin{tabular}{c||l|cc|cc|cc|cc|cc|cc}
    \toprule
     
\multirow{2}{*}{Rank} & \multirow{2}{*}{\diagbox [width=8em,trim=l] {\texttt{A}}{\texttt{B}}} & \multicolumn{2}{c|}{PerfectDou}& \multicolumn{2}{c|}{DouZero} & \multicolumn{2}{c|}{DeltaDou} & \multicolumn{2}{c|}{RHCP-v2} & \multicolumn{2}{c|}{CQN} &\multicolumn{2}{c}{Random}\\
\cline{3-14}
& & P & L & P & L & P & L & P & L & P & L & P & L \\
    \hline
    \midrule
     1 & PerfectDou (Ours) & \textbf{0.622}$^*$ & 0.378$^*$ & \textbf{0.640}$^*$ & 0.446$^*$ & \textbf{0.693}$^*$ & 0.474$^*$ & \textbf{0.609}$^*$ & 0.478$^*$ & \textbf{0.894}$^*$ & \textbf{0.830}$^*$ & \textbf{0.998}$^*$ & \textbf{0.990}$^*$\\
     2 & DouZero (Public) & \textbf{0.554}$^*$ & 0.360$^*$ & \textbf{0.584}$^*$ & 0.416$^*$ & \textbf{0.684}$^*$ & 0.487$^*$ & 0.427$^*$ & 0.475$^*$ & \textbf{0.851}$^*$ & \textbf{0.769}$^*$ & \textbf{0.992}$^*$ & \textbf{0.986}$^*$\\
     3 & DeltaDou & \textbf{0.526}$^*$ & 0.307$^*$ & \textbf{0.513}$^*$ & 0.317$^*$ & \textbf{0.588} & 0.412 & \textbf{0.768}$^*$ & \textbf{0.614}$^*$ & \textbf{0.835} & \textbf{0.733} & \textbf{0.996} & \textbf{0.987}\\
     4 & RHCP-v2 & \textbf{0.522}$^*$ & 0.391$^*$ & \textbf{0.525}$^*$ & \textbf{0.573}$^*$ & 0.386$^*$ & 0.232$^*$ & \textbf{0.536}$^*$ & 0.434$^*$ & \textbf{0.687}$^*$ & \textbf{0.853}$^*$ & \textbf{0.994}$^*$ & \textbf{0.985}$^*$\\
     5 & CQN & 0.170$^*$ & 0.106$^*$ & 0.231$^*$ & 0.149$^*$ & 0.267 & 0.165 & 0.147$^*$ & 0.313$^*$ & 0.476 & \textbf{0.524} & \textbf{0.921} & \textbf{0.857}\\
     6 & Random & 0.010$^*$ & 0.002$^*$ & 0.014$^*$ & 0.008$^*$ & 0.013 & 0.004 & 0.015$^*$ & 0.006$^*$ & 0.143 & 0.080 & \textbf{0.654} & 0.346\\
     \bottomrule
    \end{tabular}
    }
    \label{tab:complete-wp}
\end{table*}

\newpage
\section{More Implementation Details}
\label{ap:implementation}

\subsection{The Oracle for Minimum Steps to Play Out All cards}

In our paper, as mentioned in \se{sec:reward}, we utilize an oracle for evaluating the minimum steps to play out all cards. Particularly, the oracle is implemented by a dynamic programming algorithm combined with depth-first-search, which can be referred to \url{https://www.cnblogs.com/SYCstudio/p/7628971.html} (which is also a competition problem of National Olympiad in Informatics in Provinces (NOIP) 2015). For completeness, we summarize the pseudocode for implementing such an algorithm in \alg{alg:oracle}.

\begin{algorithm} 
	\caption{Calculate Minimum Step to Play Out All Cards} 
	\label{alg:oracle}
	
	\begin{algorithmic}[1]
	
	\Function{\texttt{Main}}{}
	\State $N_1 \gets$ all possible number of single card
	\State $N_2 \gets$ all possible number of pair card
	\State $N_3 \gets$ all possible number of trio card
	\State $N_4 \gets$ all possible number of bomb card
	
	\Function{\texttt{InitMatrix}}{$F$}
	\For{$action$ $\in$ \{\textbf{\textit{Solo}}, \textbf{\textit{Pair}}, \textbf{\textit{Trio}}, \textbf{\textit{Bomb}}, \textbf{\textit{Chain-of-Trio}}, \textbf{\textit{Trio-with-Pair}}, \textbf{\textit{Quad-with-Solos}}, \textbf{\textit{Quad-with-Pair}}, \textbf{\textit{Quad-with-Pairs}}\}}
	\State $d_1  \gets$ the number of \textbf{\textit{Solo}} card in $action$
	\State $d_2  \gets$ the number of \textbf{\textit{Pair}} card in $action$
	\State $d_3  \gets$ the number of \textbf{\textit{Trio}} card in $action$
	\State $d_4  \gets$ the number of \textbf{\textit{Bomb}} card in $action$
	\State $F[ N_1, N_2, N_3, N_4 ] \gets \min( F[ N_1, N_2, N_3, N_4 ], F[ N_1 - d_1, N_2 - d_2, N_3 - d_3, N_4 - d_4 ] + 1 )$
	\If{$action$ is \textbf{\textit{Trio}}}
	\State $F[ N_1, N_2, N_3, N_4 ] \gets \min( F[ N_1, N_2, N_3, N_4 ], F[ N_1 + 1, N_2 + 2, N_3 - 1, N_4 ] )$
	\EndIf
	\If{$action$ is \textbf{\textit{Bomb}}}
	\State $F[ N_1, N_2, N_3, N_4 ] \gets \min( F[ N_1, N_2, N_3, N_4 ], F[ N_1 + 1, N_2 + 2, N_3 - 1, N_4 ] )$
	\EndIf
	\EndFor
	\EndFunction
	
	\Function{\texttt{NowStep}}{$Cards$}
	\If{\textbf{\textit{Rocket}} $\in$ $Cards$}
    \State $left\_cards$ $\gets$ left cards after playing out \textbf{\textit{Rocket}}
	\State \Return $\min$( \texttt{NowStep} ( $left\_cards$ ) + 1, \texttt{NowStep} ( $left\_cards$ ) + 2 )
	\EndIf
	\If{only one \textbf{\textit{Joker}} $\in$ $Cards$}
    \State $left\_cards$ $\gets$ left cards after playing out \textbf{\textit{Joker}}
	\State \Return \texttt{NowStep} ( $left\_cards$ ) + 1
	\EndIf
	\If{no \textbf{\textit{Joker}} $\in$ $Cards$}
	\State calculate number of \textbf{\textit{Solo}} $N_1$, \textbf{\textit{Pair}} $N_2$, \textbf{\textit{Trio}} $N_3$ and \textbf{\textit{Bomb}} $N_4$ of $Cards$
	\State \Return $F[ N_1, N_2, N_3, N_4 ]$
	\EndIf
	\EndFunction
	
	\Function{\texttt{DFS}}{$step$, $ans$, $Cards$}
	\If{$step$ $>$ $ans$} 
	\State \Return $ans$
	\EndIf
	\State $ans$ $\gets \min$( $step$, \texttt{NowStep}($Cards$) )
	\For{\textbf{\textit{Chain-of-Solo}} $\in$ $Cards$}
	\State $left\_cards$ $\gets$ left cards after playing out \textbf{\textit{Chain-of-Solo}}
	\State \texttt{DFS}($step$ + 1, $ans$, $left\_cards$ of $Cards$)
	\EndFor
	\For{\textbf{\textit{Chain-of-Pair}} in $Cards$}
	\State $left\_cards$ $\gets$ left cards after playing out \textbf{\textit{Chain-of-Pair}}
	\State \texttt{DFS}($step$ + 1, $ans$, $left\_cards$ of $Cards$)
	\EndFor
	\For{\textbf{\textit{Plane-with-Solo}} $\in$ $Cards$}
	\State $left\_cards$ $\gets$ left cards after playing out \textbf{\textit{Plane-with-Solo}}
	\State \texttt{DFS}($step$ + 1, $ans$, $left\_cards$ of $Cards$)
	\EndFor
	\EndFunction
	
	\State Create a matrix $F$ of size $[N_1,N_2,N_3,N_4]$
	\State {\texttt{InitMatrix}($F$)}
    \State {$step$ $\gets$ 0 ,$ans$ $\gets +\infty$, $Cards$ $\gets$ all $Cards$ to be calculated}
    \State{\texttt{DFS}( 0, $ans$, $Cards$ )}
    \EndFunction
	\end{algorithmic} 
\end{algorithm}

\subsection{Detailed Action Space}
\label{ap:action-space}

In our paper, we utilize a simplified discrete action space of 621 for learning the actor, since we observe that the original action space of 27472 (include all possible combinations) contains a large number of actions that can be abstract.
For instance, actions like Bomb with kickers and Trio with kickers occupy the action space most due to the large number of combinations of kickers. To this end, we abstract actions with the same main cards into one action, and significantly reduce the action space.
When the policy chooses an abstract action, we further deploy a simple decoding function to obtain the most preferred action in the original action space, as illustrated in \alg{alg:decode}. We note that this part is similar to an old implementation of RLCard~\cite{zha2019rlcard}: \url{https://github.com/datamllab/rlcard/blob/d100952f144e4b0fd7186cc06e79ef277cda9722/rlcard/envs/doudizhu.py#L67}.

Below we list all the 621 discrete actions of PerfectDou in 15 categories, where the notation * in category (9) (10) (11) and (12) denotes the kicker. 
\begin{algorithm} 
	\caption{Decode action} 
	\label{alg:decode}
	
	\begin{algorithmic}[1]
	
	\Function{\texttt{Decode}}{$M$}
	\State $A \gets $get all available actions from current hand
	\State $K \gets$get all kickers using the main card $M$ from $A$
	\For{$k$ in $K$}
	\State calculate score $s$ of each $k$ 
	\State $N \gets$ number of actions contains $k$ in $A$, $rank_k \gets$ card rank of $k$ 
	\State $s \gets 1.0 * N + 0.1 * rank_k$
	\EndFor
	\State \Return $k$ with minimum $s$
    \EndFunction
	\end{algorithmic} 
\end{algorithm}

(1) Solo (15 actions): 3, 4, 5, 6, 7, 8, 9, T\footnote{T for Ten (10).}, J, Q, K, A, 2, B, R

(2) Pair (13 actions): 33, 44, 55, 66, 77, 88, 99, TT, JJ, QQ, KK, AA, 22

(3) Trio (13 actions): 333, 444, 555, 666, 777, 888, 999, TTT, JJJ, QQQ, KKK, AAA, 222

(4) Trio with Solo (182 actions): 3334, 3335, 3336, 3337, 3338, 3339, 333T, 333J, 333Q, 333K, 333A, 3332, 333B, 333R, 3444, 4445, 4446, 4447, 4448, 4449, 444T, 444J, 444Q, 444K, 444A, 4442, 444B, 444R, 3555, 4555, 5556, 5557, 5558, 5559, 555T, 555J, 555Q, 555K, 555A, 5552, 555B, 555R, 3666, 4666, 5666, 6667, 6668, 6669, 666T, 666J, 666Q, 666K, 666A, 6662, 666B, 666R, 3777, 4777, 5777, 6777, 7778, 7779, 777T, 777J, 777Q, 777K, 777A, 7772, 777B, 777R, 3888, 4888, 5888, 6888, 7888, 8889, 888T, 888J, 888Q, 888K, 888A, 8882, 888B, 888R, 3999, 4999, 5999, 6999, 7999, 8999, 999T, 999J, 999Q, 999K, 999A, 9992, 999B, 999R, 3TTT, 4TTT, 5TTT, 6TTT, 7TTT, 8TTT, 9TTT, TTTJ, TTTQ, TTTK, TTTA, TTT2, TTTB, TTTR, 3JJJ, 4JJJ, 5JJJ, 6JJJ, 7JJJ, 8JJJ, 9JJJ, TJJJ, JJJQ, JJJK, JJJA, JJJ2, JJJB, JJJR, 3QQQ, 4QQQ, 5QQQ, 6QQQ, 7QQQ, 8QQQ, 9QQQ, TQQQ, JQQQ, QQQK, QQQA, QQQ2, QQQB, QQQR, 3KKK, 4KKK, 5KKK, 6KKK, 7KKK, 8KKK, 9KKK, TKKK, JKKK, QKKK, KKKA, KKK2, KKKB, KKKR, 3AAA, 4AAA, 5AAA, 6AAA, 7AAA, 8AAA, 9AAA, TAAA, JAAA, QAAA, KAAA, AAA2, AAAB, AAAR, 3222, 4222, 5222, 6222, 7222, 8222, 9222, T222, J222, Q222, K222, A222, 222B, 222R

(5) Trio with Pair (156 actions): 33344, 33355, 33366, 33377, 33388, 33399, 333TT, 333JJ, 333QQ, 333KK, 333AA, 33322, 33444, 44455, 44466, 44477, 44488, 44499, 444TT, 444JJ, 444QQ, 444KK, 444AA, 44422, 33555, 44555, 55566, 55577, 55588, 55599, 555TT, 555JJ, 555QQ, 555KK, 555AA, 55522, 33666, 44666, 55666, 66677, 66688, 66699, 666TT, 666JJ, 666QQ, 666KK, 666AA, 66622, 33777, 44777, 55777, 66777, 77788, 77799, 777TT, 777JJ, 777QQ, 777KK, 777AA, 77722, 33888, 44888, 55888, 66888, 77888, 88899, 888TT, 888JJ, 888QQ, 888KK, 888AA, 88822, 33999, 44999, 55999, 66999, 77999, 88999, 999TT, 999JJ, 999QQ, 999KK, 999AA, 99922, 33TTT, 44TTT, 55TTT, 66TTT, 77TTT, 88TTT, 99TTT, TTTJJ, TTTQQ, TTTKK, TTTAA, TTT22, 33JJJ, 44JJJ, 55JJJ, 66JJJ, 77JJJ, 88JJJ, 99JJJ, TTJJJ, JJJQQ, JJJKK, JJJAA, JJJ22, 33QQQ, 44QQQ, 55QQQ, 66QQQ, 77QQQ, 88QQQ, 99QQQ, TTQQQ, JJQQQ, QQQKK, QQQAA, QQQ22, 33KKK, 44KKK, 55KKK, 66KKK, 77KKK, 88KKK, 99KKK, TTKKK, JJKKK, QQKKK, KKKAA, KKK22, 33AAA, 44AAA, 55AAA, 66AAA, 77AAA, 88AAA, 99AAA, TTAAA, JJAAA, QQAAA, KKAAA, AAA22, 33222, 44222, 55222, 66222, 77222, 88222, 99222, TT222, JJ222, QQ222, KK222, AA222

(6) Chain of Solo (36 actions): 34567, 45678, 56789, 6789T, 789TJ, 89TJQ, 9TJQK, TJQKA, 345678, 456789, 56789T, 6789TJ, 789TJQ, 89TJQK, 9TJQKA, 3456789, 456789T, 56789TJ, 6789TJQ, 789TJQK, 89TJQKA, 3456789T, 456789TJ, 56789TJQ, 6789TJQK, 789TJQKA, 3456789TJ, 456789TJQ, 56789TJQK, 6789TJQKA, 3456789TJQ, 456789TJQK, 56789TJQKA, 3456789TJQK, 456789TJQKA, 3456789TJQKA

(7) Chain of Pair (52 actions): 334455, 445566, 556677, 667788, 778899, 8899TT, 99TTJJ, TTJJQQ, JJQQKK, QQKKAA, 33445566, 44556677, 55667788, 66778899, 778899TT, 8899TTJJ, 99TTJJQQ, TTJJQQKK, JJQQKKAA, 3344556677, 4455667788, 5566778899, 66778899TT, 778899TTJJ, 8899TTJJQQ, 99TTJJQQKK, TTJJQQKKAA, 334455667788, 445566778899, 5566778899TT, 66778899TTJJ, 778899TTJJQQ, 8899TTJJQQKK, 99TTJJQQKKAA, 33445566778899, 445566778899TT, 5566778899TTJJ, 66778899TTJJQQ, 778899TTJJQQKK, 8899TTJJQQKKAA, 33445566778899TT, 445566778899TTJJ, 5566778899TTJJQQ, 66778899TTJJQQKK, 778899TTJJQQKKAA, 33445566778899TTJJ, 445566778899TTJJQQ, 5566778899TTJJQQKK, 66778899TTJJQQKKAA, 33445566778899TTJJQQ, 445566778899TTJJQQKK, 5566778899TTJJQQKKAA

(8) Chain of Trio (45 actions): 333444, 444555, 555666, 666777, 777888, 888999, 999TTT, TTTJJJ, JJJQQQ, QQQKKK, KKKAAA, 333444555, 444555666, 555666777, 666777888, 777888999, 888999TTT, 999TTTJJJ, TTTJJJQQQ, JJJQQQKKK, QQQKKKAAA, 333444555666, 444555666777, 555666777888, 666777888999, 777888999TTT, 888999TTTJJJ, 999TTTJJJQQQ, TTTJJJQQQKKK, JJJQQQKKKAAA, 333444555666777, 444555666777888, 555666777888999, 666777888999TTT, 777888999TTTJJJ, 888999TTTJJJQQQ, 999TTTJJJQQQKKK, TTTJJJQQQKKKAAA, 333444555666777888, 444555666777888999, 555666777888999TTT, 666777888999TTTJJJ, 777888999TTTJJJQQQ, 888999TTTJJJQQQKKK, 999TTTJJJQQQKKKAAA

(9) Plane with Solo (38 actions): 333444**, 444555**, 555666**, 666777**, 777888**, 888999**, 999TTT**, TTTJJJ**, JJJQQQ**, QQQKKK**, KKKAAA**, 333444555***, 444555666***, 555666777***, 666777888***, 777888999***, 888999TTT***, 999TTTJJJ***, TTTJJJQQQ***, JJJQQQKKK***, QQQKKKAAA***, 333444555666****, 444555666777****, 555666777888****, 666777888999****, 777888999TTT****, 888999TTTJJJ****, 999TTTJJJQQQ****, TTTJJJQQQKKK****, JJJQQQKKKAAA****, 333444555666777*****, 444555666777888*****, 555666777888999*****, 666777888999TTT*****, 777888999TTTJJJ*****, 888999TTTJJJQQQ*****, 999TTTJJJQQQKKK*****, TTTJJJQQQKKKAAA*****

(10) Plane with Pair (30 actions): 333444****, 444555****, 555666****, 666777****, 777888****, 888999****, 999TTT****, TTTJJJ****, JJJQQQ****, QQQKKK****, KKKAAA****, 333444555******, 444555666******, 555666777******, 666777888******, 777888999******, 888999TTT******, 999TTTJJJ******, TTTJJJQQQ******, JJJQQQKKK******, QQQKKKAAA******, 333444555666********, 444555666777********, 555666777888********, 666777888999********, 777888999TTT********, 888999TTTJJJ********, 999TTTJJJQQQ********, TTTJJJQQQKKK********, JJJQQQKKKAAA********

(11) Quad with Solo (13 actions): 3333**, 4444**, 5555**, 6666**, 7777**, 8888**, 9999**, TTTT**, JJJJ**, QQQQ**, KKKK**, AAAA**, 2222**

(12) Quad with Pair (13 actions): 3333****, 4444****, 5555****, 6666****, 7777****, 8888****, 9999****, TTTT****, JJJJ****, QQQQ****, KKKK****, AAAA****, 2222****

(13) Bomb (13 actions): 3333, 4444, 5555, 6666, 7777, 8888, 9999, TTTT, JJJJ, QQQQ, KKKK, AAAA, 2222

(14) Rocket (1 action): BR

(15) Pass (1 action): PASS

\end{document}